%% file: main.tex
\documentclass[letterpaper]{article}

\usepackage[preprint]{aaai2027}
\usepackage[hyphens]{url}
\usepackage{graphicx}
\usepackage{natbib}
\usepackage{caption}
\usepackage{booktabs}
\usepackage{amsfonts}
\usepackage{amsmath}
\usepackage{amssymb}
\usepackage{adjustbox}

\newcommand{\Lp}{L_{p}}
\newcommand{\Dfgsm}{D_{\mathrm{FGSM}}}
\newcommand{\Dpgd}{D_{\mathrm{PGD}}}
\newcommand{\eps}{\varepsilon}
\newcommand{\PGD}[1]{\mathrm{PGD}\text{-}#1}
\newcommand{\TableFont}{\small}

\title{Margin-Drop Coordinates for Cross-Budget Robustness Evaluation}

\author{
    Yanliang Huang\textsuperscript{\rm 1},
    Zhen Zhang\textsuperscript{\rm 1},
    Peng Xie\textsuperscript{\rm 1},
    Wenyuan Wu\textsuperscript{\rm 1},
    Sitong Zhu\textsuperscript{\rm 2},
    Zhuoqi Zeng\textsuperscript{\rm 3},
    Amr Alanwar\textsuperscript{\rm 1}
}
\affiliations{
    \textsuperscript{\rm 1}School of Computation, Information and Technology,
    Technical University of Munich, Munich, Germany\\
    \textsuperscript{\rm 2}Department of Statistics,
    Ludwig-Maximilians-Universit{\"a}t M{\"u}nchen, Munich, Germany\\
    \textsuperscript{\rm 3}School of Engineering,
    Hainan Bielefeld University of Applied Sciences, Hainan, China\\
    \{yanliang.huang, zhenzhang.zhang, p.xie, wenyuan.wu, alanwar\}@tum.de,
    sitong.zhu@campus.lmu.de,
    zhuoqi.zeng@hainan-biuh.edu.cn
}

\begin{document}

\maketitle

\begin{abstract}
Fixed-budget robustness evaluation can select the wrong frozen vision encoder.
An encoder that survives a shallow attack may lose most of that robustness when
the same evaluation is strengthened. We ask whether the shallow evaluation
contains enough information to identify this budget fragility. For each
clean-correct sample, the evaluation records the clean pairwise margin, the
first-order linearized margin-drop scale, the margin drop from a clean-start
one-step attack, and the drop reached by an iterative attack. Normalizing by
that scale gives three margin-drop coordinates capturing clean margin slack,
one-step shortfall, and drift, where drift is the additional normalized margin
drop the iterative attack reaches beyond the one-step perturbation. Together, they
reconstruct the normalized post-attack margin and therefore the pass-or-fail
outcome. Across $42$ pretrained frozen vision encoders, the shallow survival
rate carries essentially no rank information about subsequent $\PGD{10}$ to
$\PGD{200}$ collapse, at Spearman $-0.006$, while the median shallow drift
coordinate ranks the same collapse at $+0.811$. The result persists in a
held-out encoder pool and under an $\ell_\infty$ evaluation. With deep
evaluation limited to $11$ encoders, ranking by shallow drift recovers $11$ of
the $17$ high-collapse encoders, compared with $5$ under survival-rate ranking.
The full coordinate decomposition further distinguishes cases that share the
same fixed-budget residual but diverge at deeper budgets, and separates margin
repair from drift repair under interventions, revealing distinct repair paths that
endpoint robustness alone does not identify.
\end{abstract}

\input{sections/introduction}
\input{sections/formalization}
\input{sections/setup}
\input{sections/main_result}
\input{sections/intervention}
\input{sections/related_work}
\input{sections/conclusion}

\bibliography{references}

\clearpage
\appendix
\input{sections/appendix}

\end{document}

%% file: sections/introduction.tex
\section{Introduction}
\label{sec:intro}

A fixed-budget robustness evaluation can rank the wrong encoder. Its survival
rate reports performance under the stated attack, but the ordering may change
when the attack is strengthened. In our 42-encoder pool, ViT-B/16 has
$60.1\%$ survival under a 10-step PGD evaluation and $8.4\%$ under a
200-step evaluation, whereas DINO ViT-S/16 loses only $2.0$ percentage points
under the same strengthening. An encoder that appears competitive under a
shallow attack can therefore be highly fragile to evaluation budget.

Pretrained vision encoders are screened across model zoos
~\citep{goldblum2023battle} and reused as components of larger systems,
including vision-language models, vision-language-action policies, and
downstream perception pipelines
~\citep{liu2023llava,brohan2023rt2,schlarmann2024robust}. Adversarial
robustness evaluation is computationally expensive, so broad evaluations use
bounded attack configurations or reduced evaluation sets. Battle of the
Backbones uses 20-step PGD for its ImageNet adversarial-robustness
comparison~\citep{goldblum2023battle}. Robust CLIP restricts adversarial
evaluation to 1,000 samples per dataset and reports only clean performance for
its larger LLaVA-13B evaluation because of the computational cost of
adversarial robustness evaluation~\citep{schlarmann2024robust}. We write
$\PGD{k}\times r$ for a $k$-step PGD evaluation with $r$ restarts per
step-size setting. In our pool, increasing the budget from
$\PGD{10}\times5$ to $\PGD{200}\times5$ raises the recorded attack-loop time
from $2.15$ to $41.03$ GPU-hours. The practical question is whether shallow evaluation can support initial model screening when model rankings may shift as the evaluation budget increases, which affects the validity of robustness-based safety assessment.

\begin{figure*}[!t]
 \centering
 \begin{minipage}[c]{0.42\linewidth}
  \includegraphics[width=\linewidth]{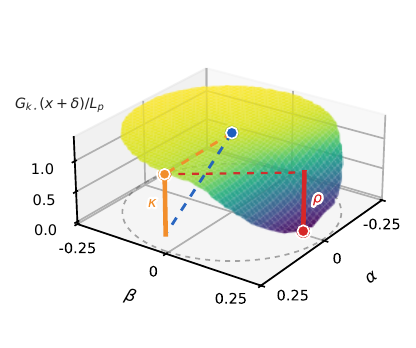}
 \end{minipage}\hspace{0.03\linewidth}
 \begin{minipage}[c]{0.40\linewidth}
  \includegraphics[width=\linewidth]{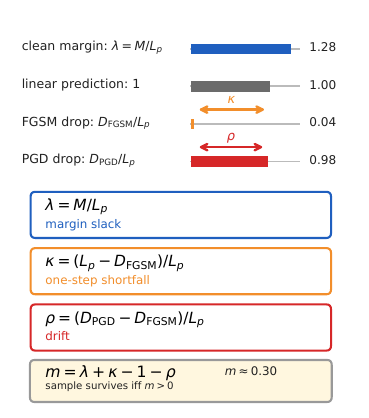}
 \end{minipage}
 \caption{\textbf{Coordinate decomposition of measured margin drop.}
 \emph{(Left)} Normalized fixed-competitor pairwise margin over a
 two-dimensional slice of the $\ell_2$ threat set for a real ViT-B/16 sample.
 The two in-plane axes are signed input-space coordinates, with $\alpha$ along
 the first-order attack direction at the clean point and $\beta$ along the
 component of the selected PGD perturbation orthogonal to it, while the surface
 height is the normalized margin. Because the two directions are orthonormal,
 the dashed circle $\alpha^2+\beta^2=\eps^2$, with $\eps=0.25$, is the
 threat-set boundary. Blue, orange, and red mark the clean input, the one-step
 perturbation, and the selected iterative perturbation. The coordinate
 $\kappa$ is the normalized shortfall of the realized one-step drop relative
 to the first-order prediction, while drift $\rho$ is the additional
 normalized drop reached by the iterative attack beyond the one-step
 perturbation. The left panel shows the geometry orthogonal to the first-order
 attack direction associated with this drift. \emph{(Right)} The same recorded quantities give the clean margin
 slack $\lambda$ and reconstruct the normalized post-attack margin
 $m=\lambda+\kappa-1-\rho$, so the sample survives the stated attack when
 $m>0$.}
 \label{fig:soul}
\end{figure*}

We find that the shallow evaluation record contains more information than its
survival rate. For each clean-correct sample, it contains four quantities, which are
the clean pairwise margin between the true class and its clean nearest competitor,
a first-order attack scale $L_p$, the margin drop reached by a clean-start
one-step perturbation, and the drop reached by the iterative attack. Here
$L_p$ is the largest margin drop predicted by the clean-point linearization
within the stated threat set. Expressing the margin quantities in units of
$L_p$ gives three margin-drop coordinates, which are margin slack $\lambda$,
one-step shortfall $\kappa$, and drift $\rho$. Drift measures the additional normalized
margin drop reached by the iterative attack beyond the one-step perturbation.
The coordinates reconstruct the normalized post-attack margin,
\[
m=\lambda+\kappa-1-\rho,
\]
and the sample survives the stated attack when $m>0$. The survival rate keeps
only the fraction of samples that satisfy this condition, while the coordinates
retain how each post-attack margin is formed. For a fixed threat set,
$\lambda$ and $\kappa$ are determined by the clean and one-step record, while
$\rho$ changes with the iterative evaluation budget. Figure~\ref{fig:soul}
illustrates the coordinates and the corresponding survival-margin ledger on a
real ViT-B/16 sample.

Across the 42 pretrained frozen vision encoders, median shallow drift
$\bar\rho_{10}$ ranks the collapse from $\PGD{10}\times5$ to
$\PGD{200}\times5$ at Spearman $+0.811$, while the shallow survival rate
carries essentially no rank information for the same target at $-0.006$.
The result holds on a held-out encoder pool, under an $\ell_\infty$
evaluation, and across restart, solver, dynamic-competitor, and family-aware
checks. With deep evaluation limited to 11 of the 42 encoders, drift ranking
recovers 11 of the 17 encoders that lose at least 20 percentage points,
compared with 5 under survival-rate ranking. The full coordinate decomposition
also distinguishes fixed-budget outcomes with different shortfall and drift
structures. In intervention studies on ViT-B/16 and Swin-B, the coordinates
separate margin repair from drift repair and reveal mechanism differences that
endpoint robustness alone does not distinguish.

Our contributions are:
\begin{itemize}
\item We introduce per-sample margin-drop coordinates for clean margin slack,
one-step shortfall, and drift on a common first-order scale. The coordinates
reconstruct fixed-budget survival and isolate drift as the coordinate that
changes with the iterative evaluation budget.

\item We evaluate shallow drift across 42 pretrained frozen vision encoders.
Median shallow drift consistently ranks cross-budget collapse across held-out
encoders, an $\ell_\infty$ evaluation, solver and restart variants, dynamic
competitor recomputation, and family-aware uncertainty analysis. Drift-based
ranking also identifies more high-collapse encoders when deep-evaluation
capacity is limited.

\item We use the full coordinate decomposition to distinguish fixed-budget
outcomes with different margin-drop structures and to evaluate repair
interventions through changes in margin slack, drift, and the full survival
margin.
\end{itemize}

%% file: sections/formalization.tex
\section{Margin-Drop Coordinates}
\label{sec:formalization}

\subsection{Evaluation record and normalization}
\label{ssec:def}

Let $f$ be a frozen encoder and $h$ a linear head producing logits
$z=h(f(x))\in\mathbb{R}^K$. A sample $(x,y)$ is clean-correct if
$\arg\max_j z_j(x)=y$. For each clean-correct sample, we fix the clean
nearest competitor
\[
k_\star=\arg\max_{j\neq y} z_j(x)
\]
and define the fixed-competitor pairwise margin
\begin{equation}
\label{eq:margin}
G_{k_\star}(u)=z_y(u)-z_{k_\star}(u).
\end{equation}
We write
\[
M=G_{k_\star}(x)
\]
for the clean margin and keep $k_\star$ fixed after perturbation, so all
budget-specific measurements refer to the same pairwise margin function.

Let
\[
g=\nabla_x G_{k_\star}(x)
\]
and let $\mathcal{B}$ denote the stated threat set. The clean-start one-step
perturbation is
\begin{equation}
\label{eq:onestep}
\delta_{\mathrm{FGSM}}\in
\arg\max_{\delta\in\mathcal{B}}
-\langle g,\delta\rangle.
\end{equation}
It maximizes the margin decrease predicted by the clean-point linearization.
For an $\ell_2$ ball,
\[
\delta_{\mathrm{FGSM}}=-\eps\frac{g}{\|g\|_2},
\]
and for an $\ell_\infty$ ball,
\[
\delta_{\mathrm{FGSM}}=-\eps\,\mathrm{sign}(g).
\]

For an iterative budget $k$, let $\mathcal{C}_k$ contain the final
perturbations returned by all stated PGD restarts and step-size settings.
We select
\begin{equation}
\label{eq:pgd_selection}
\delta_{\mathrm{PGD}}^{(k)}\in
\arg\max_{\delta\in\mathcal{C}_k\cup\{\delta_{\mathrm{FGSM}}\}}
\left[M-G_{k_\star}(x+\delta)\right].
\end{equation}
Thus the budget-$k$ record selects the largest measured margin drop among the
final PGD endpoints and the one-step candidate.

The resulting sample-level record contains four quantities:
\begin{align}
M
&=G_{k_\star}(x),\\
L_p
&=-\langle g,\delta_{\mathrm{FGSM}}\rangle, \label{eq:Lp}\\
\Dfgsm
&=M-G_{k_\star}(x+\delta_{\mathrm{FGSM}}), \label{eq:d1}\\
\Dpgd^{(k)}
&=M-G_{k_\star}(x+\delta_{\mathrm{PGD}}^{(k)}). \label{eq:dk}
\end{align}
The scale $L_p$ is the largest margin drop predicted by the clean-point
linearization over $\mathcal{B}$. For an $\ell_p$ ball of radius $\eps$,
\[
L_p=\eps\|g\|_q,
\qquad
\frac{1}{p}+\frac{1}{q}=1.
\]
We assume $L_p>0$ and exclude samples with zero or numerically zero clean
gradient.

The normalization by $L_p$ removes an arbitrary positive logit scale across
independently trained heads. For a fixed evaluation record, a positive
rescaling $z\mapsto cz$ with $c>0$ gives
\[
(M,L_p,\Dfgsm,\Dpgd^{(k)})
\mapsto
c(M,L_p,\Dfgsm,\Dpgd^{(k)}),
\]
so the normalized coordinates defined below are unchanged.

\subsection{Coordinates and fixed-budget survival}
\label{ssec:coordinates}

For budget $k$, define
\begin{align}
\lambda
&=\frac{M}{L_p}, \label{eq:lambda}\\
\kappa
&=\frac{L_p-\Dfgsm}{L_p}, \label{eq:kappa}\\
\rho_k
&=\frac{\Dpgd^{(k)}-\Dfgsm}{L_p}. \label{eq:rho}
\end{align}
Margin slack $\lambda$ measures the clean margin in units of the first-order
attack scale. One-step shortfall $\kappa$ measures the signed difference
between the first-order predicted drop and the realized one-step drop.
Positive $\kappa$ means that the one-step perturbation realizes less margin
drop than the clean-point linearization predicts, while negative $\kappa$
means that it realizes more.

We call $\rho_k$ the drift at budget $k$. It measures the additional normalized
margin drop reached by the iterative attack beyond the one-step perturbation.
Because the one-step candidate is included in
$\mathcal{C}_k\cup\{\delta_{\mathrm{FGSM}}\}$,
\[
\rho_k\ge0.
\]
On $600$ saved perturbations, $\rho_{10}$ tracks the component of the measured
gain orthogonal to the initial attack ray at Spearman $+0.982$.

The definitions give
\[
\begin{aligned}
\Dfgsm&=(1-\kappa)L_p, &
M&=\lambda L_p,\\
\Dpgd^{(k)}&=(1-\kappa+\rho_k)L_p.
\end{aligned}
\]
Therefore,
\begin{equation}
\label{eq:identity}
G_{k_\star}(x+\delta_{\mathrm{PGD}}^{(k)})
=
M-\Dpgd^{(k)}
=
L_p(\lambda+\kappa-1-\rho_k).
\end{equation}
We define the normalized post-attack survival margin
\begin{equation}
\label{eq:survival_margin}
m_k
\equiv
\lambda+\kappa-1-\rho_k.
\end{equation}
Since $L_p>0$, the sample survives the stated budget-$k$ evaluation if and
only if
\[
m_k>0.
\]

The corresponding clean-correct conditional retention is
\begin{equation}
\label{eq:retention}
R_k
=
\Pr\!\left(
m_k>0
\,\middle|\,
G_{k_\star}(x)>0
\right).
\end{equation}
Thus the survival rate retains only the fraction of samples whose post-attack
margin remains positive, while $(\lambda,\kappa,\rho_k)$ retains the
decomposition of that margin.

\subsection{Cross-budget behavior and the net residual}
\label{ssec:budget_consequence}

At a fixed threat set and radius, $\lambda$ and $\kappa$ are determined by the
clean and one-step record. Only $\rho_k$ depends on the iterative evaluation
budget. For two budgets $k<k'$ with the remaining protocol fixed, define the
cross-budget drift change
\begin{equation}
\label{eq:delta_rho}
\Delta\rho_{k\to k'}
=
\rho_{k'}-\rho_k
=
\frac{\Dpgd^{(k')}-\Dpgd^{(k)}}{L_p}.
\end{equation}
The two budget-specific evaluations may be run independently, so
$\Delta\rho_{k\to k'}$ compares two evaluation records and can take either
sign.

The survival margins satisfy
\begin{equation}
\label{eq:budget_transition}
m_{k'}
=
m_k-\Delta\rho_{k\to k'}.
\end{equation}
For a sample that survives budget $k$, so that $m_k>0$, failure at budget
$k'$ occurs exactly when
\begin{equation}
\label{eq:flip_condition}
\Delta\rho_{k\to k'}\ge m_k.
\end{equation}
The current survival margin therefore specifies how much additional
cross-budget drift change is required to cross the pairwise decision boundary.

The scalar net residual
\begin{equation}
\label{eq:tau}
\tau_k
=
\rho_k-\kappa
=
\frac{\Dpgd^{(k)}-L_p}{L_p}
\end{equation}
compresses the two directional coordinates into a single quantity. Since
\[
m_k=\lambda-1-\tau_k,
\]
the pair $(\lambda,\tau_k)$ reproduces fixed-budget survival. However, for any
$c\ge0$,
\[
(\kappa,\rho_k)
\longmapsto
(\kappa+c,\rho_k+c)
\]
leaves both $\tau_k$ and $m_k$ unchanged. For example,
\[
(\kappa,\rho_k)=(0,0)
\qquad\text{and}\qquad
(\kappa,\rho_k)=(a,a)
\]
have the same net residual for any $a\ge0$, despite having different
shortfall and drift.

The net residual therefore preserves the fixed-budget outcome while removing
the split between $\kappa$ and $\rho_k$. This split matters across budgets
because the budget change enters the survival margin through $\rho_k$.
Section~\ref{sec:main_result} tests whether shallow drift carries information
about later collapse and whether fixed-budget outcomes with similar residuals
can hide different deeper-budget behavior.

For encoder-level analyses, an overbar denotes the median over clean-correct
samples of an encoder, as in $\bar\rho_k$ and $\bar\tau_k$.

%% file: sections/setup.tex
\section{Evaluation setup}
\label{sec:setup}

Our main evaluation uses frozen pretrained vision encoders with independently
trained linear heads on ImageNet-100~\citep{deng2009imagenet}. All
margin-based quantities use the
fixed clean competitor defined in Section~\ref{sec:formalization}. The primary
$\ell_2$ protocol uses radius $\eps=0.25$ in pixel space $[0,1]$, and an
additional $\ell_\infty$ evaluation uses $\eps=0.375/255$. These radii preserve
non-degenerate shallow survival across the encoder pool, and sensitivity to the
evaluation radius is reported in Appendix~\ref{app:eps_sensitivity}.

\paragraph{Models and data.}
We evaluate 42 frozen pretrained vision encoders spanning convolutional,
transformer, self-supervised, and multimodal model families, including ResNet,
ConvNeXt, EfficientNet, DenseNet, MobileNet, Mixer, ViT, DeiT, DeiT3, Swin,
DINO, DINOv2, and CLIP. The pool follows the broad model-zoo setting considered
in prior encoder benchmarks~\citep{goldblum2023battle,robustart}.
For each encoder, we train a linear classifier on ImageNet-100 while keeping
the representation fixed. All encoders are evaluated on the same fixed
validation subset of $994$ images drawn at $10$ images per class. The primary cross-budget analysis uses all 42
encoders. Analyses that require variation away from the retention floor use a
40-encoder analytical subset obtained by excluding two floor-saturated models
whose retention at $\PGD{50}$ is below $5\%$, and these
analyses are marked explicitly. Appendix~\ref{app:pool} lists model
families, checkpoints, clean accuracies, and per-encoder records, and
Appendix~\ref{app:in1k} reports an ImageNet-1K transfer evaluation with retrained 1000-way linear
heads, including a 42-encoder shallow evaluation and an 8-encoder deeper-budget
study.

\paragraph{Attack protocol.}
The primary shallow evaluation is $\PGD{10}\times5$, and
$\PGD{200}\times5$ serves as the deep reference. We write
$\PGD{k}\times r$ for a $k$-step PGD evaluation with $r$ restarts for each
step-size setting in the step-size set $\{\eps/4,\eps/10\}$. For each budget, we pool the
final endpoints returned by all restart and step-size configurations together
with the clean-start one-step perturbation $\delta_{\mathrm{FGSM}}$, and select
the candidate with the largest fixed-competitor margin drop. The resulting
budget-specific quantity is $\Dpgd^{(k)}$.

We additionally evaluate a lower-cost $\PGD{10}\times1$ variant and use
$\PGD{50}\times10$ for sensitivity analyses and intervention measurements.
For the primary $\ell_2$ protocol, each PGD step is projected onto the
$\ell_2$ threat set and the perturbed input is clipped to $[0,1]$.
The $\ell_\infty$ projection, step-size choices, restart conventions, and
additional $\PGD{500}$ evaluations are specified in
Appendices~\ref{app:convergence} and~\ref{app:linf}, and
Appendix~\ref{app:untargeted_ce} reports the alignment checks that use standard
untargeted cross-entropy attacks.

\paragraph{Primary statistic and cross-budget target.}
For every clean-correct encoder--sample pair, we record
$M$, $\Lp$, $\Dfgsm$, and the budget-specific $\Dpgd^{(k)}$, and derive
$(\lambda,\kappa,\rho_k)$ using
Equations~\ref{eq:lambda}--\ref{eq:rho}. For encoder $e$, the primary shallow
statistic is the median drift
\[
\bar\rho_{10}^{(e)}
=
\operatorname{median}_{i\in\mathcal I_e}\rho_{10,i},
\]
where $\mathcal I_e$ contains the clean-correct samples for that encoder.

Let $R_k^{(e)}$ denote the clean-correct conditional pairwise retention defined
in Section~\ref{ssec:budget_consequence}. The primary cross-budget target is
the percentage-point retention loss, which we call cross-budget collapse,
\[
\Delta R^{(e)}_{10\to200}
=
R_{10}^{(e)}-R_{200}^{(e)}.
\]
The main analysis asks whether $\bar\rho_{10}$ ranks
$\Delta R_{10\to200}$ across the 42 encoders. Secondary evaluations vary the
restart count, attack solver, threat norm, competitor definition, encoder
pool, and evaluation radius while preserving the same margin-drop
construction.

%% file: sections/main_result.tex
\section{Shallow drift ranks cross-budget collapse}
\label{sec:main_result}

\subsection{Cross-budget ranking}
\label{ssec:rho_eval}

\begin{figure*}[!t]
 \centering
 \begin{minipage}[t]{0.43\linewidth}
  \includegraphics[width=\linewidth]{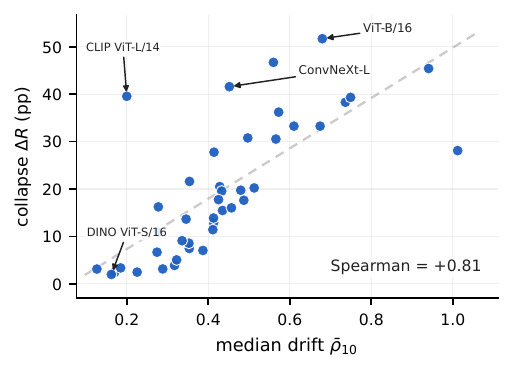}
 \end{minipage}\hspace{0.03\linewidth}
 \begin{minipage}[t]{0.43\linewidth}
  \includegraphics[width=\linewidth]{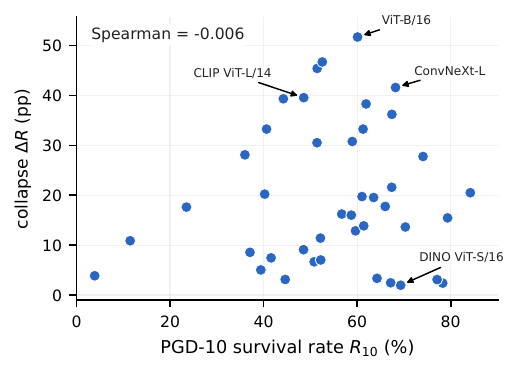}
 \end{minipage}
 \caption{\textbf{Shallow-budget drift ranks deeper-budget collapse.}
 \emph{(Left)} Across the 42-encoder $\ell_2$ evaluation,
 $\bar\rho_{10}$ ranks
 $\Delta R(\PGD{10}\times5\to\PGD{200}\times5)$ at Spearman $+0.811$.
 \emph{(Right)} The shallow survival rate $R_{10}$ has essentially no rank
 relationship with the same collapse target.}
 \label{fig:rho_collapse_scatter}
\end{figure*}

Across the 42-encoder pool, the encoder-level median shallow drift ranks the
collapse from $\PGD{10}\times5$ to $\PGD{200}\times5$ at
\[
\begin{aligned}
&\mathrm{Spearman}\!\left(
\bar\rho_{10},
\Delta R(\PGD{10}\times5\to\PGD{200}\times5)
\right)\\
&\hspace{7em}=+0.811.
\end{aligned}
\]
Figure~\ref{fig:rho_collapse_scatter} contrasts this relationship with the
shallow survival rate $R_{10}$, whose Spearman correlation with the same
collapse target is $-0.006$. The normalization matters as much as the
quantity, since on the same full-pool record the raw margin-drop gap ranks
collapse at $+0.580$ and normalization by the clean margin gives $+0.488$, against
$+0.811$ for normalization by $L_p$.

Budget-transition measurements connect shallow drift to the cross-budget
relation in Equation~\ref{eq:budget_transition}. On a 20-encoder transition
subset, $\bar\rho_{10}$ ranks the encoder-level median of
$\rho_{200}-\rho_{10}$ at Spearman $+0.884$ and ranks
$\bar\rho_{200}$ at $+0.914$. Encoders with larger shallow drift therefore
tend to exhibit larger cross-budget drift changes as the evaluation budget
increases,
which directly reduces the survival margin through
Equation~\ref{eq:budget_transition}.

The ranking remains strong across changes in encoder pool, threat norm, attack
solver, restart count, and competitor definition. Table~\ref{tab:validation_matrix}
summarizes the main checks. The held-out encoder pool gives Spearman $+0.882$,
the $\ell_\infty$ evaluation gives $+0.819$, and the family-cluster bootstrap
interval for the primary $\ell_2$ correlation is $[+0.423,+0.945]$.
The dynamic all-class check re-selects the competitor at each perturbed input
and gives $+0.797$.

\begin{table*}[!t]
 \centering
 \TableFont
 \caption{Validation of shallow drift ranking. Correlation rows report
 Spearman statistics. The family-cluster row reports a bootstrap interval for
 the primary correlation.}
 \label{tab:validation_matrix}
 \begin{tabular}{lll}
    \toprule
    Check & Target & Result \\
    \midrule
    Primary evaluation
      & $\Delta R(\PGD{10}\times5\to\PGD{200}\times5)$
      & $+0.811$ \\
    One-restart evaluation
      & $\Delta R(\PGD{10}\times1\to\PGD{200}\times5)$
      & $+0.717$ \\
    Conditional collapse
      & $\Pr(\mathrm{fail}_{200}\mid\mathrm{pass}_{10})$
      & $+0.669$ \\
    Family-cluster bootstrap
      & 95\% CI for primary Spearman
      & $[+0.423,+0.945]$ \\
    Dynamic all-class margin
      & $\Delta R_{\mathrm{all}}(\PGD{10}\times5\to\PGD{200}\times5)$
      & $+0.797$ \\
    Cross-entropy Auto-PGD top-$1$
      & $\Delta R_{\mathrm{top1}}^{\mathrm{CE/APGD}}(10\to100)$
      & $+0.833$ \\
    $\ell_\infty$ evaluation
      & $\Delta R^{\ell_\infty}(\PGD{10}\times5\to\PGD{100}\times5)$
      & $+0.819$ \\
    Held-out encoder pool, $n=11$
      & $\Delta R(\PGD{10}\times5\to\PGD{200}\times5)$
      & $+0.882$ \\
    ImageNet-1K, $n=8$
      & $\Delta R(\PGD{10}\to\PGD{200})$
      & $+0.905$ \\
    Shallow survival baseline
      & $R_{10}$ vs.\ primary collapse
      & $-0.006$ \\
    Net-residual baseline
      & $\bar\tau_{10}$ vs.\ primary collapse
      & $-0.013$ \\
    \bottomrule
 \end{tabular}
\end{table*}

\subsection{Residual compression and cancellation}
\label{sec:decomposition}

\begin{figure*}[!t]
 \centering
 \begin{minipage}[t]{0.43\linewidth}
  \includegraphics[width=\linewidth]{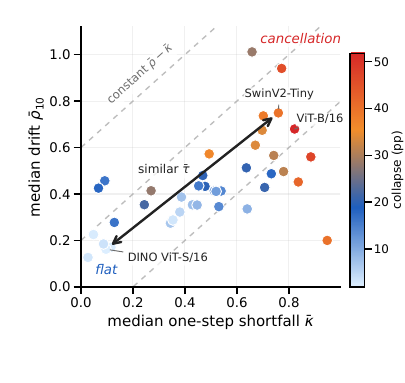}
 \end{minipage}\hspace{0.03\linewidth}
 \begin{minipage}[t]{0.43\linewidth}
  \includegraphics[width=\linewidth]{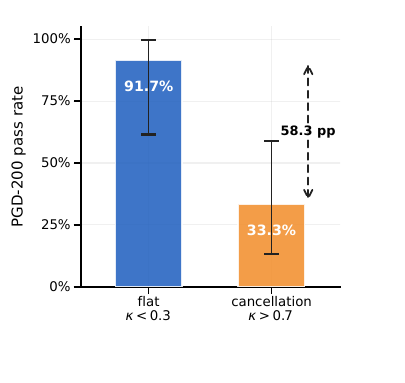}
 \end{minipage}
 \caption{\textbf{Residual compression hides distinct shortfall--drift
 structure.}
 \emph{(Left)} Encoder-level coordinates on the $n=40$ analytical subset show
 that similar net residuals can correspond to different drift and
 cross-budget collapse.
 \emph{(Right)} In a DeiT3 ViT-B/16 band matched on $\tau_{10}$ and
 $\lambda$, flat and cancellation samples have sharply different
 $\PGD{200}$ survival.}
 \label{fig:decomposition_necessity}
\end{figure*}

At the encoder level, the net residual itself has essentially no rank
relationship with cross-budget collapse, at Spearman $-0.013$ in
Table~\ref{tab:validation_matrix}. The net residual $\tau_k=\rho_k-\kappa$ can
hide distinct shortfall--drift structures with the same fixed-budget outcome. A flat sample
has low $\kappa$ and low $\rho$, whereas a cancellation sample has high
$\kappa$ and high $\rho$. Figure~\ref{fig:decomposition_necessity} shows that
these regimes can have similar net residuals but very different cross-budget
behavior.

Matched encoder pairs show the distinction at the model level.
DINO ViT-S/16 and SwinV2-Tiny differ in $\bar\tau_{10}$ by only $0.02$, but
their $\bar\rho_{10}$ values differ by $0.59$ and their cross-budget collapse
differs by $37.4$\,pp. DeiT-Base and EfficientNet-B0 show the same pattern,
with a $34.9$\,pp collapse difference. The complete matched-pair analysis is
reported in Appendix~\ref{app:matched_pairs}.

The distinction also remains after matching samples on their current
fixed-budget state. Within a DeiT3 ViT-B/16 band satisfying
$\tau_{10}\in[-0.1,+0.1]$ and $\lambda\in[1.0,4.0]$, flat samples pass
$\PGD{200}$ at $91.7\%$, compared with $33.3\%$ for cancellation samples.
The same ordering holds across alternative $\lambda$ bands. In a five-encoder diagnostic subset spanning
$\bar\rho_{50}\in[0.19,1.87]$, among samples that already survive $\PGD{200}$, a further $\PGD{500}$ evaluation punctures
$27.9\%$ of the cancellation regime, compared with at most $9.9\%$ in the
other regimes. The shortfall--drift structure compressed by $\tau$ therefore
corresponds to different behavior as the evaluation budget is increased.

\subsection{Prioritizing models for deeper evaluation}
\label{ssec:prioritize}

A practical use of shallow drift is to prioritize which models receive deeper
evaluation. Suppose a model pool has already been evaluated with the shallow
PGD budget, but only a subset can subsequently be evaluated with the deep
budget. The models most important to identify are those for which the shallow
evaluation is misleading, namely those that later undergo large cross-budget
collapse. In our 42-encoder pool, 17 encoders lose at least 20 percentage
points when the evaluation is strengthened from $\PGD{10}\times5$ to
$\PGD{200}\times5$. If 11 encoders can be sent to the deeper evaluation,
selecting the 11 with largest shallow drift identifies 11 of these 17
budget-fragile encoders, whereas selecting 11 by shallow survival rate
identifies 5.

The final comparison among the selected candidates uses the deep-budget
results, while shallow drift determines which candidates reach that budget.
Across five fragility thresholds from $10$ to $30$ percentage points and
five deep-evaluation
capacities from $5$ to $21$ encoders, drift ranking recovers more
budget-fragile encoders than survival-rate ranking in all $70$ combinations
spanned by three deep budgets, and survival-rate ranking falls below the
uniform random expectation in $54$ of them. Appendix~\ref{app:selection}
reports the full grid.

%% file: sections/intervention.tex
\section{Evaluating survival-margin repair with coordinates}
\label{sec:intervention}

The cross-budget analysis uses shallow drift to identify encoders whose
robustness deteriorates as the evaluation budget increases. We now use the full
margin-drop decomposition to measure how robustness interventions change the
survival margin. For
$c\in\{\lambda,\kappa,\rho,m\}$, we summarize the change from baseline to an
intervention on matched samples as
\[
\Delta\bar c
=
\operatorname{median}_i
\left(
c_i^{\mathrm{after}}-c_i^{\mathrm{before}}
\right).
\]
All quantities within a comparison are measured at the same PGD budget.
Because the median is nonlinear, we measure $\Delta\bar m$ directly alongside
the component medians. Table~\ref{tab:case_studies}
reports $\Delta\bar\lambda$, $\Delta\bar\rho$, and $\Delta\bar m$.

\subsection{Intervention setup}
\label{ssec:coord_objective}

We compare two LoRA-based interventions under matched rank, optimizer, batch
size, training budget, and inner-attack budget. The first applies
local-linearity regularization (LLR)~\citep{qin2019adversarial}. The second,
denoted Coord, uses the coordinate-targeted objective
\[
\mathcal{L}
=
\mathcal{L}_{\mathrm{clean}}
+
\alpha[\tilde\rho_\theta]_+^2
+
\beta[\kappa_0-\eta_\kappa-\kappa_\theta]_+^2
+
\gamma[\lambda_0-\eta_\lambda-\lambda_\theta]_+^2 .
\]
Here $\mathcal{L}_{\mathrm{clean}}$ is cross-entropy on clean inputs,
$[a]_+=\max(a,0)$, the subscript $0$ denotes quantities measured on the
frozen baseline, and $\theta$ denotes the LoRA-adapted model evaluated against
the same fixed competitor. The drift term penalizes a signed training-time
estimate of drift, while the remaining guard terms limit reductions in $\kappa$ and
$\lambda$ relative to the baseline. We compute
\[
\tilde\rho_\theta
=
\frac{
D_{\mathrm{PGD},\theta}
-
D_{\mathrm{FGSM},\theta}
}{
L_{p,0}
},
\]
using the frozen baseline scale $L_{p,0}$. This keeps the normalization fixed
while the student parameters are updated. The tilde distinguishes the signed
training-time quantity from the nonnegative evaluation drift defined by the
candidate-pooling rule in Section~\ref{sec:formalization}. Both methods use a
10-step inner attack. Appendix~\ref{app:intervention} provides the training
procedure, hyperparameters, normalization details, guard-term ablations, and
head-retraining experiments.

\subsection{Intervention case studies}
\label{ssec:vit_b16_closure}

\paragraph{ViT-B/16.}
Coord yields the larger improvement in $R_{50}$, and the coordinate changes
attribute this ordering to the joint movement of drift and margin slack. LLR
reduces drift more than Coord, while gaining almost no margin slack,
$\Delta\bar\lambda=+0.079$ against $+0.400$ for Coord, and it produces the
smaller survival-margin change as a result.

ViT-B/16 also provides a deeper cross-budget check. On the intervention-specific matched split, the baseline has $R_{10}=61.2\%$ and
$R_{200}=8.3\%$, corresponding to a $52.9$\,pp collapse.
LLR reduces this collapse by $14.0$\,pp to $38.9\pm1.9$\,pp, while Coord
reduces it by $17.7$\,pp to $35.2\pm0.7$\,pp. The Coord-minus-LLR difference
in remaining collapse is $-3.73$\,pp, with paired-bootstrap interval
$[-5.91,-1.58]$\,pp. The deeper evaluation therefore preserves the ordering
seen in the PGD-50 survival-margin changes.

\begin{table}[!t]
 \centering
 \TableFont
 \caption{Intervention case studies through survival-margin coordinates.
 All robustness and coordinate changes are measured at $\PGD{50}\times10$ and
 reported as four-seed means. $\Delta R_{50}$ is the change in pairwise robust
 retention relative to the matched baseline.}
 \label{tab:case_studies}

 \begin{adjustbox}{max width=\linewidth}
 \begin{tabular}{llrrrr}
    \toprule
    Encoder & Method
    & $\Delta R_{50}$
    & $\Delta\bar\lambda$
    & $\Delta\bar\rho_{50}$
    & $\Delta\bar m$ \\
    \midrule
    ViT-B/16 & LLR   & $+27.9$\,pp & $+0.079$ & $-0.491$ & $+0.485$ \\
    ViT-B/16 & Coord & $+33.3$\,pp & $+0.400$ & $-0.337$ & $+0.687$ \\
    Swin-B   & LLR   & $+7.5$\,pp  & $-0.601$ & $-1.015$ & $+0.466$ \\
    Swin-B   & Coord & $+25.7$\,pp & $+0.003$ & $-0.541$ & $+0.666$ \\
    \bottomrule
 \end{tabular}
 \end{adjustbox}
\end{table}

\paragraph{Swin-B.}
\label{ssec:swin_h2h}
Swin-B shows the same trade-off in a sharper form. LLR again produces the
larger drift reduction, but here it erodes margin slack outright, with
$\Delta\bar\lambda=-0.601$, while Coord holds margin slack at its baseline
value. Coord therefore produces both the larger survival-margin change and the
larger improvement in $R_{50}$.

\paragraph{Architecture-matched robust encoders.}
We also compare four robust encoders with architecture-matched standard
counterparts based on ResNet-18, ConvNeXt-B, Swin-B, and CLIP ViT-L/14.
Across the four matched pairs, totaling $3{,}432$ common clean-correct
samples, all four robust variants have
zero collapse from $\PGD{10}\times5$ to $\PGD{200}\times5$ at the primary
radius. Their dominant coordinate movement is increased margin slack, where
$\Delta\bar\lambda$ ranges from $+4.27$ to $+39.18$, while
$\Delta\bar\rho$ ranges from $-0.93$ to $+0.01$. The shared mechanism is
therefore substantially greater survival slack, accompanied by
method-dependent drift changes.

In the two LoRA case studies, LLR primarily improves robustness by reducing drift through local-linearity regularization, while Coord combines drift reduction with explicit preservation of margin slack. The decomposition shows whether improved budget stability is associated with greater margin slack, lower drift, or both.

%% file: sections/related_work.tex
\section{Related work}
\label{sec:related}

Adversarial robustness evaluation depends critically on attack adequacy.
Weak attacks, gradient obfuscation, and poorly chosen attack configurations
can substantially overestimate robustness
~\citep{uesato2018adversarial,athalye2018obfuscated,carlini2019evaluating}.
AutoAttack, RobustBench, and AttackBench improve evaluation reliability through
complementary attacks, standardized protocols, and systematic comparisons of
attack implementations
~\citep{croce2020reliable,croce2021robustbench,attackbench}.
Evaluation efficiency has also motivated methods such as Minimum-Margin Attack,
which seeks reliable robustness estimates at substantially lower computational
cost~\citep{gao2022minimummargin}.
Together, these works show that reported robustness depends on both the
strength of the evaluation and the protocol used to obtain it.

Margins and local geometry have been widely used to characterize robustness at
the sample level. Local linearity and curvature have been used as robustness
regularizers~\citep{qin2019adversarial,moosavi2019cure} and to study
instability in single-step adversarial training
~\citep{andriushchenko2020understanding,wong2020fast,dejorge2022make}.
Sample-level robustness measures and predictors use logit margins, local
Lipschitz estimates, randomized smoothing, margin consistency, or learned
attackability signals to quantify vulnerability or certified robustness
~\citep{weng2018clever,tsuzuku2018lipschitz,cohen2019smoothing,ngnawe2024margin,raina2023identifying}.
Our margin-drop decomposition uses quantities already produced by a shallow
evaluation to characterize cross-budget fragility.

At the model level, large-scale robustness studies compare architectures,
pretraining strategies, adaptation methods, and adversarially trained models
through endpoint performance
~\citep{robustart,goldblum2023battle,shao2022adversarial,bai2021transformers,schlarmann2024robust,mao2023understanding,cagatan2025ssl,salman2020adversarially}.
Our evaluation focuses on how these model comparisons change as the attack
budget is increased. We use the decomposition to compare how robustness
interventions change margin slack, drift, and the resulting survival margin.

%% file: sections/conclusion.tex
\section{Conclusion}
\label{sec:conclusion}

Fixed-budget robustness rankings can change substantially as the attack budget
increases, and shallow survival rate alone does not reveal which encoders are
budget-fragile. Our margin-drop decomposition retains three components of the
same evaluation record, which are margin slack, one-step shortfall, and drift,
with drift as the budget-dependent coordinate. Across 42 pretrained frozen vision
encoders, shallow drift consistently ranks later cross-budget collapse, with
the same pattern persisting across held-out encoders, an $\ell_\infty$
evaluation, and multiple attack and uncertainty checks. The ranking also
prioritizes high-collapse encoders more effectively than shallow survival when
deep-evaluation capacity is limited. The full decomposition further
distinguishes fixed-budget outcomes with different shortfall and drift
structures, and in intervention studies reveals whether robustness
improvements arise from margin repair, drift repair, or changes in both.
Extending the analysis to a broader pool of adversarially trained encoders
and to end-to-end multimodal systems would test whether the same
drift-based cross-budget structure persists under different margin geometry
and attack-convergence regimes.

%% file: sections/appendix.tex
\section*{Appendix organization}
\newcommand{\AppTableFont}{\TableFont}

Appendix~\ref{app:pool} specifies the encoder pool and evaluation records.
Appendix~\ref{app:identity} gives notation, the identity derivation,
coordinate conventions, and the norm-general form. Appendix~\ref{app:convergence}
reports PGD restart provenance, budget-transition checks, one-budget predictor baselines,
deep-reference comparisons, and selection sensitivity. Appendix~\ref{app:threshold}
contains held-out encoder validation, cross-radius checks, cross-norm
construction notes, and ImageNet-1K class-scale validation. Appendix~\ref{app:matched_pairs}
reports directional-split details, matched pairs, and lambda-band robustness.
Appendix~\ref{app:offray_geometry} gives perturbation-vector geometry, and
Appendix~\ref{app:sample_flip} gives sample-level flip prediction.
Appendix~\ref{app:intervention} contains intervention ablations and closure
checks. Appendix~\ref{app:additional_validation} reports dynamic-margin,
solver-variant, and head-seed checks.
Appendix~\ref{app:untargeted_ce} reports the standard top-$1$ CE/APGD check.
Appendix~\ref{app:nc} gives representation-geometry context.
Appendix~\ref{app:linf} reports the $\ell_\infty$ evaluation, including
cross-norm rank consistency, a radius sweep, the cancellation regime,
and attack adequacy.

\section{Encoder pool and evaluation records}
\label{app:pool}

The $42$-encoder primary pool spans convolutional, transformer,
self-supervised, and multimodal encoders. The analytical $n = 40$ pool excludes
CLIP-RN50 and CLIP-RN101, whose $R_{50}$ values are below $5\%$.
The encoder pool draws backbones from ResNet \citep{he2016deep}, ConvNeXt
\citep{liu2022convnet}, EfficientNet \citep{tan2019efficientnet}, DenseNet
\citep{huang2017densely}, MobileNetV3 \citep{howard2019searching},
MLP-Mixer \citep{tolstikhin2021mlp}, DeiT \citep{touvron2021training} and
DeiT3 \citep{touvron2022deit}, Swin \citep{liu2021swin}, supervised ViT
\citep{dosovitskiy2020image}, DINO \citep{caron2021emerging}, DINOv2
\citep{oquab2024dinov2}, and CLIP \citep{radford2021learning}.
Table~\ref{tab:pool_summary} groups the encoders by family. Detailed
per-encoder values of $\bar\rho_{10}$, $\bar\kappa$,
$\bar\lambda$, $R_{10}$, $R_{50}$, and $R_{200}$ are provided in the
accompanying code and data artifact.

\begin{table*}[!t]
 \centering
 \AppTableFont
 \caption{Primary $42$-encoder pool grouped by architecture family. The two encoders excluded from the analytical pool are marked $\dagger$.}
 \label{tab:pool_summary}
 \begin{adjustbox}{max width=\linewidth}
 \begin{tabular}{lll}
    \toprule
    Family & Encoders & Coverage \\
    \midrule
    ResNet, timm & RN18, RN34, RN50, RN101, RN152, Wide-RN50 & supervised IN-1K \\
    ConvNeXt & ConvNeXt-T/S/B/L & supervised IN-1K and IN-22K \\
    EfficientNet & EfficientNet-B0/B1/B3 & supervised IN-1K \\
    DenseNet, MobileNet, Mixer & DenseNet-121/201, MobileNetV3-L, Mixer-B/16 & supervised IN-1K \\
    DeiT and DeiT3 & DeiT-T/S/B, DeiT3-S/M/B/L & supervised IN-1K \\
    Swin and SwinV2 & Swin-T/S/B, SwinV2-T & supervised IN-1K and IN-22K \\
    Supervised ViT & ViT-T/S/B & supervised IN-1K and IN-21K \\
    DINO & DINO ViT-S/16, DINO ViT-B/8, DINO ViT-B/16, DINO RN50 & SSL IN-1K \\
    DINOv2 & DINOv2 ViT-S/14, DINOv2 ViT-B/14, DINOv2 ViT-L/14 & SSL LVD-142M \\
    CLIP & CLIP ViT-B/16, CLIP ViT-L/14 & multimodal LAION and WIT \\
    Floor saturated & CLIP-RN50$^\dagger$, CLIP-RN101$^\dagger$ & $R_{50} < 5\%$ \\
    \bottomrule
  \end{tabular}
 \end{adjustbox}
\end{table*}

\subsection{Evaluation sampling protocol}
\label{app:sampling}

All encoders are evaluated on one fixed set of ImageNet-100 validation images,
drawn with seed $42$ at $10$ images per class over the $100$ classes, which
gives $994$ images after the class-balanced draw. The same image set is used
for every encoder and every budget, so encoder comparisons are paired at the
sample level. Coordinates are recorded on the clean-correct subset for each
encoder, which ranges from $851$ to $975$ images with median $932$, and the
packaged primary record therefore contains $39{,}172$ encoder--sample rows.

Linear heads are trained on frozen features with cross-entropy, AdamW,
learning rate $10^{-3}$, weight decay $0$, batch size $1024$, $100$ epochs,
and seed $42$. Head-seed variability is reported in
Appendix~\ref{app:head_seed}.

\subsection{Runs per reported result}
\label{app:runs}

Each encoder--budget cell is one evaluation pass, which is deterministic given
the seed and already pools $r$ restarts for each of the two step sizes together
with the one-step candidate. Encoder-level statistics are medians over the
clean-correct samples of that encoder, and reported intervals come from
family-cluster or paired bootstrap resampling. The intervention cells are the
exception and report four-seed means, and the linear-head check in
Appendix~\ref{app:head_seed}
retrains three head seeds on a $12$-backbone subset.

\subsection{Hyperparameter selection}
\label{app:hyperparams}

On the evaluation side, the quantities varied during development are the
threat radius, the restart count, the step-size set, and the PGD budget.
The $\ell_2$ cross-radius grid is $\eps\in\{0.10,0.15,0.20,0.30,0.40\}$ and
the $\ell_\infty$ radius sweep is $\eps\in\{0.25,0.375,0.5,0.75,1.0\}/255$.
The reported operating points, $\ell_2$ $\eps=0.25$ and $\ell_\infty$
$\eps=0.375/255$, were chosen because they preserve non-degenerate shallow
survival across the pool, since a saturated pool compresses the collapse
target. The $\eps=0.30$ to $\eps=0.40$ pair in Appendix~\ref{app:eps_sensitivity}
shows that saturation directly. Restart counts were tried at $1$, $5$, and
$10$, with $5$ retained for the primary shallow and deep budgets so that the
two ends of the transition match in restart count, and $10$ used at the
analytical $\PGD{50}$ budget. The step-size set is fixed at
$\{\eps/4,\eps/10\}$ for every budget. Budgets were run at
$k\in\{5,10,20,50,100,200,500\}$, and $\PGD{200}$ was retained as the deep
reference because the $\PGD{500}$ collapse ranking agrees with it at Spearman
$+0.996$. Sensitivity to each of these choices is reported in
Appendices~\ref{app:convergence}, \ref{app:threshold}, and~\ref{app:linf}.

On the intervention side, the LoRA rank and the learning rate are held fixed
across both arms so that the comparison isolates the training objective. Rank
was additionally explored at $2$, $8$, and $16$ in preliminary runs of a
different adapter objective, and rank $4$ was retained for the reported
comparison. For the coordinate-targeted objective, the training budget
fraction was tried at $0.005$ and $0.05$, the inner-attack step count at $5$
and $10$, and the guard weights at $\alpha\in\{0,0.1\}$ and
$\beta\in\{0,1\}$, where the zero settings are the unguarded ablation reported
in Appendix~\ref{app:coord_ablation}. The retained setting is the one that
matches both arms on capacity, optimizer, training budget, and inner-attack
budget.

\subsection{Computing infrastructure}
\label{app:infrastructure}

Evaluations ran on Linux compute nodes with a single data-center NVIDIA GPU per
job, either an H100 or an H200, with at least $80$\,GB of device memory. The
software environment is Python $3.13$ with PyTorch $2.11.0$ built against CUDA
$12.6$, torchvision $0.26.0$, timm $1.0.26$, open\_clip\_torch $3.3.0$,
transformers $4.57.1$, NumPy $1.26.4$, SciPy $1.17.1$, scikit-learn $1.8.0$,
pandas $2.3.3$, and Matplotlib $3.10.8$. A pinned dependency list is included
in the accompanying code and data artifact. Per-encoder cost for the shallow-to-deep
evaluation ranges from $2.15$ to $41.03$ GPU-hours.

\section{Identity derivation and coordinate conventions}
\label{app:identity}

Table~\ref{tab:notation_appendix} summarizes the notation used in the coordinate evaluation.

\begin{table*}[!t]
 \centering
 \AppTableFont
 \caption{Notation used in the coordinate evaluation. Bars denote encoder-level medians over clean-correct samples. Budget subscripts refer to the stated PGD record and its restart count.}
 \label{tab:notation_appendix}
 \begin{adjustbox}{max width=\linewidth}
 \begin{tabular}{ll}
    \toprule
    Symbol & Meaning \\
    \midrule
    $G_{k_\star}$ & fixed-competitor pairwise margin $z_y-z_{k_\star}$ \\
    $M$ & clean margin $G_{k_\star}(x)$ \\
    $\Lp$ & support-function attack scale, $\eps\|g\|_2$ for $\ell_2$ and $\eps\|g\|_1$ for $\ell_\infty$ \\
    $\Dfgsm,\Dpgd^{(k)}$ & measured one-step and budget-$k$ margin drops \\
    $\lambda$ & margin slack ratio $M/\Lp$ \\
    $\kappa$ & one-step shortfall $(\Lp-\Dfgsm)/\Lp$ \\
    $\rho_k$ & drift, the additional normalized margin drop beyond the one-step perturbation at PGD budget $k$ \\
    $\Delta\rho_{k\to k'}$ & cross-budget drift change $\rho_{k'}-\rho_k$ \\
    $\tau_k$ & net residual $\rho_k-\kappa$ \\
    $m_k$ & survival margin $\lambda+\kappa-1-\rho_k$ \\
    $R_k$ & robust retention at PGD budget $k$ \\
    $\Delta R(k_1\to k_2)$ & retention drop $R_{k_1}-R_{k_2}$ \\
    \bottomrule
  \end{tabular}
 \end{adjustbox}
\end{table*}

In the primary $\ell_2$ evaluation, the one-step perturbation is $\delta_{\mathrm{FGSM}} = -\eps g / \|g\|_2$, with $g = \nabla_x G_{k_\star}(x)$. An attack reduces $G_{k_\star}$, so the perturbation moves opposite to $g$. The realized drops $\Dfgsm$ and $\Dpgd^{(k)}$ are signed quantities. The identity holds for any $\Dfgsm, \Dpgd^{(k)} \in \mathbb{R}$. The candidate-pooling convention defined in Section~\ref{ssec:def} ensures $\Dpgd^{(k)} \ge \Dfgsm$, so the drift coordinate $\rho$ defined in Section~\ref{ssec:coordinates} is nonnegative even though the underlying margin drops are signed.

With $\Lp$ defined as the support-function scale, the survival-margin identity
follows by substitution:
\begin{align*}
G_{k_\star}(x + \delta_{\mathrm{PGD}}^{(k)}) &= M - \Dpgd^{(k)} \\
&= M - \Dfgsm - (\Dpgd^{(k)} - \Dfgsm) \\
&= M - \Lp + (\Lp - \Dfgsm) \\
&\quad - (\Dpgd^{(k)} - \Dfgsm) \\
&= \Lp\,(\lambda - 1 + \kappa - \rho) \\
&= \Lp\,(\lambda + \kappa - 1 - \rho).
\end{align*}
Since $\Lp > 0$ on every retained sample, $G_{k_\star}(x + \delta_{\mathrm{PGD}}^{(k)}) > 0$ if and only if $m > 0$.

\paragraph{Support-function form.}
Here $\delta_{\mathrm{FGSM}}$ denotes the clean-start one-step linear maximizer
defined in Section~\ref{ssec:def} for the stated threat set.
Section~\ref{ssec:def} defines the coordinates through the support-function
scale $\Lp=-\langle g,\delta_{\mathrm{FGSM}}\rangle$. The explicit form for
general $\ell_p$ threat sets is given below for reference.
For a threat set $\mathcal{B}$, define
\[
L_{\mathcal{B}}=\max_{\delta\in\mathcal{B}}-\langle g,\delta\rangle
\]
and take a one-step linear attack perturbation
$\delta_{\mathrm{FGSM}}\in\arg\max_{\delta\in\mathcal{B}}-\langle g,\delta\rangle$.
With $\Dfgsm=M-G_{k_\star}(x+\delta_{\mathrm{FGSM}})$ and
$\Dpgd^{(k)}=M-G_{k_\star}(x+\delta_{\mathrm{PGD}}^{(k)})$ for an iterative
perturbation $\delta_{\mathrm{PGD}}^{(k)}$, the definitions
\[
\lambda=\frac{M}{L_{\mathcal{B}}},
\qquad
\kappa=\frac{L_{\mathcal{B}}-\Dfgsm}{L_{\mathcal{B}}},
\]
\[
\rho_k=\frac{\Dpgd^{(k)}-\Dfgsm}{L_{\mathcal{B}}}
\]
give the same survival decomposition whenever $L_{\mathcal{B}}>0$.
For an $\ell_p$ ball of radius $\eps$, $L_{\mathcal{B}}=\eps\|g\|_q$, where
$q$ is the dual exponent satisfying $1/p+1/q=1$.

\paragraph{Domain.}
The coordinates are defined when $\Lp > 0$. For the $\ell_2$ and
$\ell_\infty$ evaluations, this is equivalent to $\|g\|_2 > 0$. We restrict the
evaluation to clean-correct samples, $M > 0$, and exclude samples with $\|g\|_2$
below floating-point precision. The packaged primary coordinate record
contains $39{,}172$ clean-correct rows after this exclusion.

\paragraph{Signed $\kappa$.}
The term $\Lp - \Dfgsm$ keeps its sign. Substituting $|\Lp - \Dfgsm|$ would break the identity when $\Dfgsm > \Lp$. We observe $\kappa < 0$ on $2{,}741$ samples and $\kappa > 1$ on $0$ samples in the packaged $39{,}172$-row primary coordinate record. The signed definition continues to satisfy the identity on those samples.

\paragraph{Finite-step curvature sign.}
A second-order Taylor expansion of $G_{k_\star}$ at $x$ along $\delta_{\mathrm{FGSM}}$ gives
\[
\Dfgsm = \Lp - \tfrac{1}{2}\delta_{\mathrm{FGSM}}^\top H_M\delta_{\mathrm{FGSM}} + O(\eps^3),
\]
where $H_M = \nabla_x^2 G_{k_\star}(x)$ is the input-space Hessian of the pairwise margin at $x$. The expansion gives a sign interpretation. Positive $\kappa$ means the second-order term offsets the first-order margin decrease along the attack ray, so the realized FGSM drop falls below the linear prediction. At the primary $\ell_2$ radius $\eps = 0.25$, $\kappa$ is a finite-step sign diagnostic.

\paragraph{Head dependence.}
For a linear head, the input-space Hessian expands as
\begin{equation}
\label{eq:HM}
H_M = \sum_i (w_{y,i} - w_{k_\star,i}) H_{f_i},
\end{equation}
where $w_{y,i}$ and $w_{k_\star,i}$ are the head weights for the true and competitor classes, and $H_{f_i}$ is the input-space Hessian of the $i$-th encoder feature. The head determines the pairwise weights and the choice of $k_\star$.

Other protocol axes also enter the recorded coordinates, where the radius
rescales $\Lp$, and PGD step and restart counts determine the reported iterative
perturbation, hence $\rho$.

\section{PGD budget checks and deep-reference comparison}
\label{app:convergence}

By construction, $\kappa$ uses the clean gradient and the one-step
perturbation, so it does not depend on the PGD budget.

A $20$-encoder diagnostic pool is used for restart and budget-transition
checks at $5$ restarts, matching the primary shallow budget in the main evaluation. We use
$\PGD{10}$ as the primary shallow budget because it matches the standard
PGD-$10$ evaluation convention while keeping cost low.
A restart sensitivity check at $\PGD{10}$ gives Spearman $> 0.97$ between $\bar\rho_{10}$ rankings at one restart and five restarts on the diagnostic pool.

The same diagnostic pool gives the budget-transition check used in the
cross-budget analysis of Section~\ref{ssec:rho_eval}. On this $20$-encoder pool,
Spearman$(\bar\rho_{10},\bar\rho_{200})=+0.914$ and
Spearman$(\bar\rho_{10},\overline{\Delta\rho}_{10\to200})=+0.884$,
where $\overline{\Delta\rho}_{10\to200}$ is the encoder-level median of
$\rho_{200}-\rho_{10}$.

\paragraph{Step-size and restart convention.}
For a budget written as $\PGD{k}\times r$, the $r$ restarts are applied to each
step size in the stated step-size set. In the primary evaluation the step-size set
is $\{\eps/4,\eps/10\}$, so $\PGD{10}\times5$ pools $10$ PGD trajectories plus
the one-step perturbation.
The first $\eps/4$ trajectory starts at zero.
All other trajectories use random starts inside the $\ell_2$ ball.
The reported PGD perturbation is the candidate with the largest measured margin
drop among all resulting trajectories and the one-step perturbation.

\subsection{Restart provenance for shallow drift}
\label{app:restart_provenance}

The primary $\bar\rho_{10}$ result uses $\PGD{10}\times5$, matched in restart count to the $\PGD{200}\times5$ deep reference.
The one-restart record is a lower-cost variant.

\begin{table}[!t]
 \centering
 \caption{Restart provenance for the shallow drift statistic. Each row reports the rank correlation for one restart setting against the stated target.}
\label{tab:restart_provenance}
\AppTableFont
\begin{adjustbox}{max width=\linewidth}
\begin{tabular}{llrr}
\toprule
Statistic & Target & $n$ & Spearman \\
\midrule
$\bar\rho_{10\times1}$ & $\Delta R(\PGD{10}\times1\to\PGD{200}\times5)$ & $42$ & $+0.717$ \\
$R_{10\times1}$ & $\Delta R(\PGD{10}\times1\to\PGD{200}\times5)$ & $42$ & $+0.031$ \\
$\bar\rho_{10\times5}$ & $\Delta R(\PGD{10}\times5\to\PGD{200}\times5)$ & $42$ & $+0.811$ \\
$R_{10\times5}$ & $\Delta R(\PGD{10}\times5\to\PGD{200}\times5)$ & $42$ & $-0.006$ \\
\bottomrule
\end{tabular}
\end{adjustbox}
\end{table}

\subsection{One-budget predictor baselines}
\label{app:predictor_baselines}

Table~\ref{tab:baselines} reports the one-budget predictor baselines used to
contextualize $\bar\rho_{10}$.

\begin{table*}[!t]
 \centering
 \AppTableFont
 \caption{One-budget predictors of $\Delta R(\PGD{10}\times5\to\PGD{200}\times5)$. $\bar\rho_{50}$ gives the gain from a deeper analytical budget on the $n = 40$ subset where the $\PGD{50}$ record is present.}
 \label{tab:baselines}
 \begin{adjustbox}{max width=\linewidth}
 \begin{tabular}{lrrrl}
    \toprule
    Predictor & $n$ & Spearman & ROC-AUC at $\ge 20$\,pp & Required record \\
    \midrule
    $\bar\rho_{50}$        & $40$ & $+0.867$ & $0.941$ & $\PGD{50}\times10$ + FGSM margins \\
    $\bar\rho_{10}$        & $42$ & $+0.811$ & $0.894$ & $\PGD{10}\times5$ + FGSM margins \\
    median raw $(\Dpgd^{(10)} - \Dfgsm)$ & $42$ & $+0.580$ & $0.821$ & $\PGD{10}\times5$ + FGSM margins \\
    $\bar\tau_{10}$        & $42$ & $-0.013$ & $0.525$ & $\PGD{10}\times5$ + FGSM margins \\
    $\bar\lambda$          & $42$ & $-0.020$ & $0.511$ & clean margin + gradient scale \\
    $R_{10}$               & $42$ & $-0.006$ & $0.480$ & $\PGD{10}\times5$ survival rate \\
    \bottomrule
  \end{tabular}
 \end{adjustbox}
\end{table*}

\subsection{Deep-reference comparison and per-sample puncture}
\label{app:reference}

We extend the evaluation to $\PGD{500}\times5$ on the full $42$-encoder pool.
The $\PGD{10}\times5\to\PGD{200}\times5$ and
$\PGD{10}\times5\to\PGD{500}\times5$ collapse rankings agree at Spearman
$+0.996$, which supports $\PGD{200}$ as the deep reference used in the
primary analysis. The primary target remains $\PGD{200}$.

For the per-sample puncture analysis, we use a five-encoder diagnostic subset
selected to span the encoder-level drift range, $\bar\rho_{50}\in[0.19,1.87]$,
where deeper-budget disagreement is most likely. The subset is DINO ViT-S/16,
DeiT3 ViT-B/16, ConvNeXt-L, ViT-B/16, and Swin-B. For each sample passing $\PGD{200}$ on this subset, we bin by
$(\kappa_{200}, \rho_{200})$ and report
$\Pr(\mathrm{fail}_{500} \mid \mathrm{pass}_{200})$ per corner bin. We label
the four corner bins by which coordinate is high or low, namely flat
($\kappa < 0.3$, $\rho < 0.3$, neither shortfall nor drift realized),
shortfall only ($\kappa > 0.7$, $\rho < 0.3$, large one-step
shortfall but little drift), drift only ($\kappa < 0.3$,
$\rho > 0.7$, little one-step shortfall but large drift), and
cancellation ($\kappa > 0.7$, $\rho > 0.7$, both effects large). The
four corner bins cover $25\%$ to $52\%$ of clean-correct samples per encoder.
Samples whose $(\kappa,\rho)$ falls in the middle band on either coordinate
are omitted from this corner-bin summary.

\begin{table*}[!t]
 \centering
 \caption{Deep-budget puncture rate by corner bin. Samples whose $(\kappa,\rho)$ falls in the middle band on either coordinate are omitted from this summary.}
 \label{tab:corner_bins}
\AppTableFont
\begin{adjustbox}{max width=\linewidth}
\begin{tabular}{lcccc}
\toprule
Bin & $n$ pass$_{200}$ & punctured & rate & $95\%$ CI \\
\midrule
flat, $\kappa < 0.3$, $\rho < 0.3$ & $478$ & $0$ & $0.00\%$ & $[0.00,0.77]$ \\
shortfall only, $\kappa > 0.7$, $\rho < 0.3$ & $81$ & $8$ & $9.88\%$ & $[4.36,18.54]$ \\
drift only, $\kappa < 0.3$, $\rho > 0.7$ & $138$ & $10$ & $7.25\%$ & $[3.53,12.92]$ \\
\textbf{cancellation, $\kappa > 0.7$, $\rho > 0.7$} & $283$ & $79$ & $\mathbf{27.92\%}$ & $\mathbf{[22.77,33.53]}$ \\
\bottomrule
\end{tabular}
\end{adjustbox}
\end{table*}

The cancellation bin's CI is disjoint from each of the other three.
Cancellation samples sit near $\lambda + \kappa \approx 1 + \rho$, where
additional PGD drop under the deeper budget is most likely to cross the
survival boundary.

\subsection{Selection sensitivity}
\label{app:selection}

Section~\ref{ssec:prioritize} reports one prioritization operating point, which sends the $11$
encoders with the largest $\bar\rho_{10}$ to the deep evaluation and recovers
$11$ of the $17$ encoders that lose at least $20$ percentage points. We vary the
fragility threshold and the deep-evaluation capacity together, and we compare
against ranking by the shallow survival rate $R_{10}$, which sends the
highest-scoring encoders forward. Writing $\mathcal{F}$ for the set of encoders
whose collapse reaches the threshold, each cell of Table~\ref{tab:selection}
reports how many members of $\mathcal{F}$ drift ranking recovers, how many
survival-rate ranking recovers, and the expectation $c|\mathcal{F}|/n$ under
uniform random selection at capacity $c$.

\begin{table*}[!t]
 \centering
 \AppTableFont
 \caption{Selection sensitivity against the $\PGD{10}\times5\to\PGD{200}\times5$
 collapse target on the $42$-encoder pool. Each cell reports drift ranking,
 survival-rate ranking, and the uniform random expectation, in that order. The
 threshold row $20$\,pp and capacity column $11$ is the operating point reported
 in Section~\ref{ssec:prioritize}.}
 \label{tab:selection}
 \begin{tabular}{rrrrrrr}
    \toprule
    & & \multicolumn{5}{c}{Deep-evaluation capacity $c$} \\
    \cmidrule(lr){3-7}
    Threshold & $|\mathcal{F}|$ & $5$ & $8$ & $11$ & $15$ & $21$ \\
    \midrule
    $10$\,pp & $29$ & $5$ / $3$ / $3.5$ & $8$ / $5$ / $5.5$ & $11$ / $7$ / $7.6$ & $15$ / $10$ / $10.4$ & $21$ / $16$ / $14.5$ \\
    $15$\,pp & $24$ & $5$ / $3$ / $2.9$ & $8$ / $4$ / $4.6$ & $11$ / $6$ / $6.3$ & $15$ / $9$ / $8.6$ & $21$ / $13$ / $12.0$ \\
    $20$\,pp & $17$ & $5$ / $2$ / $2.0$ & $8$ / $3$ / $3.2$ & $11$ / $5$ / $4.5$ & $12$ / $6$ / $6.1$ & $15$ / $9$ / $8.5$ \\
    $25$\,pp & $14$ & $5$ / $1$ / $1.7$ & $8$ / $2$ / $2.7$ & $10$ / $3$ / $3.7$ & $11$ / $4$ / $5.0$ & $13$ / $7$ / $7.0$ \\
    $30$\,pp & $12$ & $4$ / $0$ / $1.4$ & $7$ / $1$ / $2.3$ & $9$ / $2$ / $3.1$ & $10$ / $3$ / $4.3$ & $11$ / $6$ / $6.0$ \\
    \bottomrule
 \end{tabular}
\end{table*}

Drift ranking recovers more budget-fragile encoders than survival-rate ranking
in every cell. Survival-rate ranking stays near the random-selection expectation
throughout and falls below it in $54$ of the $70$ cells spanned by the three
deep budgets, which is consistent with the encoder-level Spearman of $-0.006$
between $R_{10}$ and the collapse target.

Substituting a different deep budget preserves the ordering. Repeating the grid
against $\PGD{50}\times10$ on the $n = 40$ pool where that record is present,
and against $\PGD{100}$ on the $n = 20$ diagnostic pool, leaves no cell
in which survival-rate ranking recovers as many budget-fragile encoders as
drift ranking, so the $70$ cells contain no exception.

\section{Protocol-variant and class-scale checks}
\label{app:threshold}

\paragraph{Family-cluster bootstrap.}
Family clusters are defined by architecture and pretraining family, which are
CLIP / multimodal, ConvNeXt, EfficientNet / DenseNet / MobileNet / Mixer, ResNet /
Wide-ResNet, self-supervised DINO / DINOv2, supervised ViT / DeiT /
DeiT3, and Swin / SwinV2. The interval in
Table~\ref{tab:validation_matrix} uses $10{,}000$ bootstrap resamples. Each
resample draws families with replacement, then draws encoders within each
selected family with replacement, and recomputes the primary Spearman
statistic. We report the percentile $95\%$ interval.

\subsection{Held-out encoder validation}
\label{app:heldout}

We evaluate a held-out encoder pool with the same ImageNet-100, $\ell_2$, $\eps=0.25$, $\PGD{10}\times5$, and $\PGD{200}\times5$ protocol used in the primary evaluation.
All $12$ held-out linear heads train successfully.
The reported ranking pool contains $11$ held-out encoders with $\PGD{200}$
records.

\begin{table}[!t]
 \centering
 \caption{Held-out encoder validation checks under the primary $\ell_2$ protocol.}
\label{tab:heldout_checks}
\AppTableFont
\begin{adjustbox}{max width=\linewidth}
\begin{tabular}{lcc}
\toprule
Check & Statistic & Result \\
\midrule
Collapse ranking & Spearman & $+0.882$, $p=3.30{\times}10^{-4}$ \\
Bootstrap interval & 95\% CI & $[+0.479,+1.000]$ \\
Conditional collapse & Spearman & $+0.855$, $p=8.07{\times}10^{-4}$ \\
\bottomrule
\end{tabular}
\end{adjustbox}
\end{table}

The conditional target is $\Pr(\mathrm{fail}_{200}\mid\mathrm{pass}_{10})$.
The held-out pool serves as a ranking-transfer check.

\subsection{Cross-radius sensitivity}
\label{app:eps_sensitivity}

The primary $\ell_2$ cross-$\eps$ analysis uses the analytical $n = 40$ pool and $\PGD{50}\times10$ evaluation at $\eps \in \{0.10,0.15,0.20,0.30,0.40\}$. For each pair $\eps_0 < \eps_1$, the flip rate is $\Pr(\mathrm{fail}_{\eps_1} \mid \mathrm{pass}_{\eps_0})$. We rank encoders by $\bar\rho_{50}(\eps_0)$ and compare against that flip rate.

The $\eps_0 = 0.10$ to $\eps_1 = 0.15$ pair gives Spearman $+0.727$, $p < 10^{-4}$. Controlling for $\bar\lambda(\eps_0)$ gives partial Spearman $+0.726$. Across the ten increasing pairs from the five-radius grid, nine reach Spearman $\ge +0.5$ with $p < 0.05$ and the mean is $+0.615$. The $\eps = 0.30$ to $\eps = 0.40$ pair is saturated because most encoders fail at the larger radius. The cross-$\eps$ extrapolation uses the $\PGD{50}$ analytical budget.

\subsection{Cross-norm coordinate construction}
\label{app:cross_norm}

The coordinate construction extends to general $\ell_p$ threat models through
the support-function form in Appendix~\ref{app:identity}.
Appendix~\ref{app:linf} reports the full $\ell_\infty$ evaluation.

\subsection{ImageNet-1K class-scale validation}
\label{app:in1k}

We extend the class-scale validation to ImageNet-1K through two complementary checks. First,
a full $42$-encoder ImageNet-1K $\PGD{10}\times5$ coordinate record after
training $1000$-way linear heads gives ranking transfer Spearman $+0.954$ for
$\bar\rho_{10}$ and $+0.934$ for $R_{10}$. Second, an $8$-encoder ImageNet-1K
$\PGD{200}$ record gives
Spearman$(\bar\rho_{10},\Delta R)=+0.905$, providing a class-scale check of the
shallow drift at the deeper budget under the larger class set. The
$\PGD{200}$ pilot records are reported for these encoders.

\begin{table*}[!t]
 \centering
 \caption{ImageNet-100 and ImageNet-1K records for the encoders in the class-scale check.}
 \label{tab:in1k_per_encoder}
\AppTableFont
\begin{adjustbox}{max width=\linewidth}
\begin{tabular}{lrrrrr}
\toprule
Encoder & $\bar\rho_{10}^{\mathrm{IN100}}$ & $\Delta R^{\mathrm{IN100}}$ & $\bar\rho_{10}^{\mathrm{IN1K}}$ & $\Delta R^{\mathrm{IN1K}}$ & $n$ IN-1K \\
\midrule
DINO ViT-S/16 & $+0.16$ & $+2.0$\,pp & $+0.18$ & $+2.8$\,pp & $3589$ \\
EfficientNet-B0 & $+0.18$ & $+3.4$\,pp & $+0.16$ & $+1.9$\,pp & $3800$ \\
DINOv2 ViT-S/14 & $+0.41$ & $+11.4$\,pp & $+0.39$ & $+6.8$\,pp & $3934$ \\
DeiT3 ViT-B/16 & $+0.43$ & $+15.5$\,pp & $+0.39$ & $+13.2$\,pp & $4185$ \\
CLIP ViT-B/16 & $+0.49$ & $+17.6$\,pp & $+0.45$ & $+6.6$\,pp & $3741$ \\
ConvNeXt-T & $+0.57$ & $+30.5$\,pp & $+0.51$ & $+15.2$\,pp & $4064$ \\
ViT-B/16 & $+0.68$ & $+51.7$\,pp & $+0.52$ & $+35.3$\,pp & $4214$ \\
Swin-B & $+0.94$ & $+45.4$\,pp & $+0.53$ & $+35.3$\,pp & $4098$ \\
\bottomrule
\end{tabular}
\end{adjustbox}
\end{table*}

\section{Directional-split details}
\label{app:matched_pairs}

\begin{figure}[!t]
 \centering
 \includegraphics[width=0.75\linewidth]{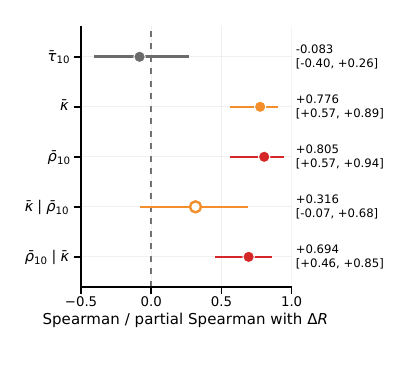}
 \caption{Marginal and partial rank statistics on the analytical $n = 40$ subset.
 Points show Spearman or partial-Spearman correlations with
 $\Delta R(\PGD{10}\times5\to\PGD{200}\times5)$. Bars show 95\% bootstrap
 confidence intervals. Rows of the form $a\mid b$ report the partial rank
 correlation for $a$ after controlling for $b$. The same values are tabulated in
 Table~\ref{tab:directional_rank}.}
 \label{fig:directional_forest}
\end{figure}

Table~\ref{tab:directional_rank} reports the marginal and partial rank
statistics behind the scalar-residual comparison. Figure~\ref{fig:directional_forest}
visualizes the same statistics as a forest plot.

\begin{table}[!t]
 \centering
 \AppTableFont
 \caption{Directional rank statistics for $\Delta R(\PGD{10}\times5\to\PGD{200}\times5)$ on the $n = 40$ analytical subset. Marginally, $\bar\tau_{10}$ has no rank signal while $\bar\kappa$ and $\bar\rho_{10}$ both rank collapse. Partial ranks concentrate the encoder-level signal in $\bar\rho_{10}$.}
 \label{tab:directional_rank}
 \begin{adjustbox}{max width=\linewidth}
 \begin{tabular}{llr}
    \toprule
    Statistic & Conditioning & Rank correlation [95\% CI] \\
    \midrule
    $\bar\tau_{10}$ & marginal & $-0.083$ $[-0.40,+0.26]$ \\
    $\bar\kappa$ & marginal & $+0.776$ $[+0.57,+0.89]$ \\
    $\bar\rho_{10}$ & marginal & $+0.805$ $[+0.57,+0.94]$ \\
    $\bar\kappa$ & controlled for $\bar\rho_{10}$ & $+0.316$ $[-0.07,+0.68]$ \\
    $\bar\rho_{10}$ & controlled for $\bar\kappa$ & $+0.694$ $[+0.46,+0.85]$ \\
    \bottomrule
  \end{tabular}
 \end{adjustbox}
\end{table}

Table~\ref{tab:matched_pairs} reports the encoder pairs satisfying $|\Delta\bar\tau_{10}| < 0.05$ and $|\Delta\bar\rho_{10}| > 0.5$ on the analytical $n = 40$ pool.

\begin{table}[!t]
 \centering
 \caption{Encoder pairs matched on the net residual, satisfying $|\Delta\bar\tau_{10}| < 0.05$ and $|\Delta\bar\rho_{10}| > 0.5$ on the analytical $n = 40$ pool. Pairs share encoders, so the table is a within-pool summary.}
 \label{tab:matched_pairs}

\AppTableFont
\begin{adjustbox}{max width=\linewidth}
\begin{tabular}{llrrr}
\toprule
Encoder A & Encoder B & $\Delta\bar\tau_{10}$ & $\Delta\bar\rho_{10}$ & $\Delta\Delta R$ \\
\midrule
DINO ViT-S/16 & SwinV2-Tiny & $-0.023$ & $-0.587$ & $-37.4$\,pp \\
DINO ViT-B/16 & SwinV2-Tiny & $-0.033$ & $-0.579$ & $-36.9$\,pp \\
DINO ViT-B/8 & SwinV2-Tiny & $+0.014$ & $-0.622$ & $-36.2$\,pp \\
EfficientNet-B0 & SwinV2-Tiny & $+0.020$ & $-0.564$ & $-36.0$\,pp \\
DeiT-Base & EfficientNet-B1 & $-0.040$ & $+0.511$ & $+35.8$\,pp \\
DeiT-Base & DINO ViT-B/8 & $+0.024$ & $+0.610$ & $+35.2$\,pp \\
DeiT-Base & EfficientNet-B0 & $+0.018$ & $+0.552$ & $+34.9$\,pp \\
DINO ViT-S/16 & Swin-Tiny & $-0.046$ & $-0.512$ & $-31.3$\,pp \\
DINO ViT-B/8 & Swin-Tiny & $-0.009$ & $-0.548$ & $-30.1$\,pp \\
Mixer-B/16 & Swin-B & $-0.019$ & $-0.515$ & $-27.7$\,pp \\
\bottomrule
\end{tabular}
\end{adjustbox}
\end{table}

\subsection{Robustness across lambda bands}
\label{app:e3_robustness}

Section~\ref{sec:decomposition} reports the $\tau_{10} \in [-0.1,+0.1]$ band on DeiT3 ViT-B/16. Alternative $\lambda$ bands keep the same ordering.

\begin{table}[!t]
 \centering
 \caption{Flat and cancellation survival across alternative $\lambda$ bands on DeiT3 ViT-B/16.}
\label{tab:lambda_bands}
\AppTableFont
\begin{adjustbox}{max width=\linewidth}
\begin{tabular}{cccc}
\toprule
$\lambda$ band & flat $n$, pass$_{200}$ & cancel $n$, pass$_{200}$ & gap \\
\midrule
$[1.0,4.0]$ & $12$, $91.7\%$ & $18$, $33.3\%$ & $+58.3$\,pp \\
$[1.5,3.5]$ & $9$, $100.0\%$ & $11$, $36.4\%$ & $+63.6$\,pp \\
$[1.0,3.0]$ & $10$, $90.0\%$ & $11$, $18.2\%$ & $+71.8$\,pp \\
\bottomrule
\end{tabular}
\end{adjustbox}
\end{table}

Within the $\tau$-matched band, high $\kappa$ also implies high $\rho$ because $\rho=\tau+\kappa$. In this band, the split indexes cancellation.

\section{Perturbation-vector geometry of drift}
\label{app:offray_geometry}

The drift terminology defined in Section~\ref{ssec:coordinates} reflects how the selected PGD
perturbation moves relative to the initial gradient ray.
We provide sample-level support through a perturbation-vector analysis.
We saved perturbation vectors for six representative encoders and $100$ clean-correct samples per encoder.
For each sample, the saved quantities are the initial attack ray $u_1=-g/\|g\|_2$, the selected $\PGD{10}\times5$ perturbation, and the selected $\PGD{200}\times5$ perturbation.
We report derived geometry summaries below. The accompanying code and data artifact contains the
summary records used here.

\begin{figure}[!t]
 \centering
 \includegraphics[width=\linewidth]{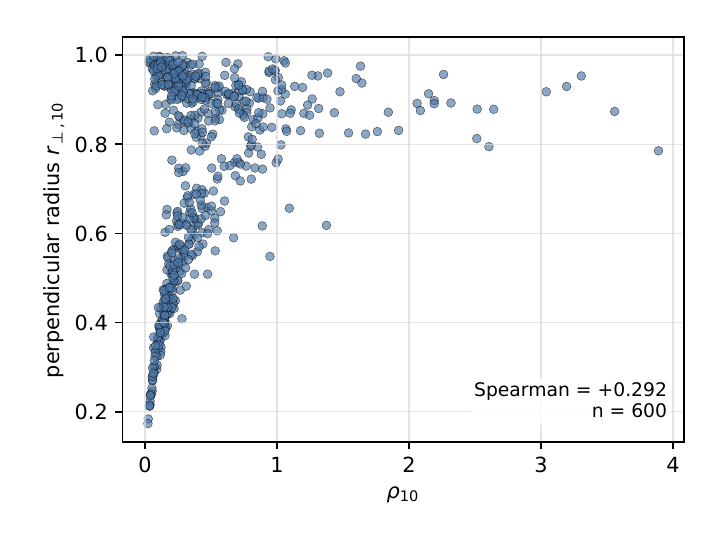}\\
 \includegraphics[width=\linewidth]{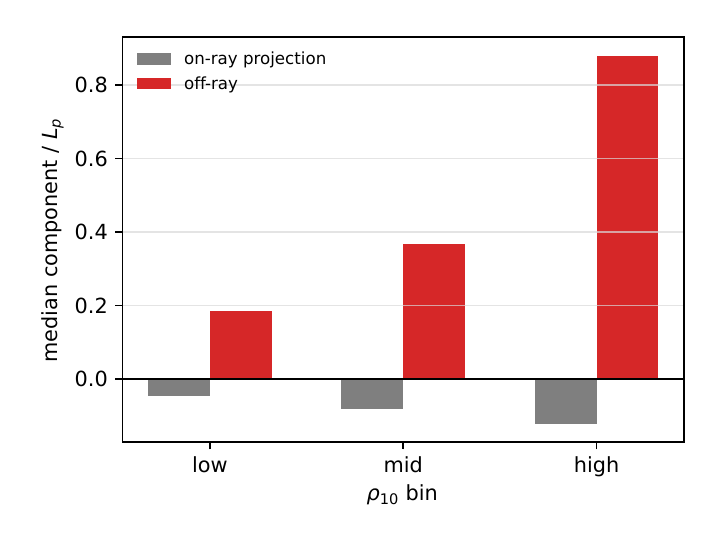}
 \caption{Perturbation-vector geometry analysis on six representative encoders and
 $600$ clean-correct samples.
 \emph{(Top)} Selected PGD perturbations are projected into components
 parallel and perpendicular to the initial gradient ray.
 \emph{(Bottom)} Decomposing the PGD-over-one-step margin drop by
 $\rho_{10}$ tertiles shows that the off-ray component accounts for most of the
 increase, while the on-ray projection component can be negative because the
 projection is not the on-ray maximizer.}
 \label{fig:offray_geometry_appendix}
\end{figure}

For a selected PGD perturbation $\delta_{\mathrm{PGD}}^{(k)}$, define
\[
\alpha=\frac{\langle \delta_{\mathrm{PGD}}^{(k)},u_1\rangle}{\eps},
\qquad
r_\perp=\frac{\|\delta_{\mathrm{PGD}}^{(k)}-\eps\alpha u_1\|_2}{\eps}.
\]
We also evaluate the pairwise margin at the on-ray projection
$\delta_{\mathrm{on}\alpha}=\eps\alpha u_1$, define
\[
D_{\mathrm{on}\alpha}=M-G_{k_\star}(x+\delta_{\mathrm{on}\alpha}),
\]
and decompose the extra PGD-over-one-step drop as
\[
D_{\mathrm{PGD}}-D_{\mathrm{FGSM}}
=
\bigl(D_{\mathrm{on}\alpha}-D_{\mathrm{FGSM}}\bigr)
+
\bigl(D_{\mathrm{PGD}}-D_{\mathrm{on}\alpha}\bigr).
\]
The first term is the on-ray component of the decomposition.
It can in principle be negative because $\delta_{\mathrm{on}\alpha}$ is the on-ray
projection of the PGD perturbation, not the on-ray maximizer.
The second term is the off-ray contribution.
Only the total $D_{\mathrm{PGD}}-D_{\mathrm{FGSM}}$ is guaranteed nonnegative
by the candidate-set convention.
Figure~\ref{fig:offray_geometry_appendix} summarizes the saved-vector geometry
and gain-decomposition results.

\begin{table}[!t]
 \centering
 \caption{Rank statistics for the saved perturbation-vector geometry analysis.}
\label{tab:offray_rank}
\AppTableFont
\begin{adjustbox}{max width=\linewidth}
\begin{tabular}{lrrr}
\toprule
Statistic & Spearman & $p$ & $n$ \\
\midrule
$\rho_{10}$ vs. off-ray gain / $L_p$ & $+0.982$ & $<10^{-12}$ & $600$ \\
\bottomrule
\end{tabular}
\end{adjustbox}
\end{table}

The off-ray gain component is the dominant contributor to $\rho_{10}$ in this
saved-vector analysis, supporting the survival-margin role of the drift
coordinate.

Binning samples by $\rho_{10}$ shows that both perpendicular displacement and
off-ray gain increase across the drift coordinate.
The median normalized on-ray components in the low, mid, and high $\rho_{10}$
bins are $-0.046$, $-0.081$, and $-0.122$, respectively.

\begin{table}[!t]
 \centering
 \caption{Perpendicular displacement and off-ray gain by $\rho_{10}$ tertile.}
\label{tab:offray_bins}
\AppTableFont
\begin{adjustbox}{max width=\linewidth}
\begin{tabular}{lrrr}
\toprule
$\rho_{10}$ bin & median $\rho_{10}$ & median $r_{\perp,10}$ & median off-ray gain / $L_p$ \\
\midrule
low & $0.139$ & $0.463$ & $0.185$ \\
mid & $0.290$ & $0.849$ & $0.367$ \\
high & $0.748$ & $0.878$ & $0.880$ \\
\bottomrule
\end{tabular}
\end{adjustbox}
\end{table}

Among shallow survivors, samples later punctured by $\PGD{200}$ already show
larger $\rho_{10}$, larger perpendicular movement, and larger off-ray gain at
the shallow budget.

\begin{table*}[!t]
 \centering
 \caption{Saved-vector geometry for shallow survivors, grouped by their deep-budget outcome.}
 \label{tab:offray_survivors}
\AppTableFont
\begin{adjustbox}{max width=\linewidth}
\begin{tabular}{lrrrr}
\toprule
Group & $n$ & median $\rho_{10}$ & median $r_{\perp,10}$ & median off-ray gain / $L_p$ \\
\midrule
pass$_{10}$ and pass$_{200}$ & $236$ & $0.219$ & $0.563$ & $0.308$ \\
pass$_{10}$ and fail$_{200}$ & $142$ & $0.396$ & $0.930$ & $0.439$ \\
fail$_{10}$ & $222$ & $0.347$ & $0.852$ & $0.459$ \\
\bottomrule
\end{tabular}
\end{adjustbox}
\end{table*}

We treat this analysis as sample-level geometry context for the drift coordinate.

\section{Per-sample flip prediction}
\label{app:sample_flip}

\begin{figure}[!t]
 \centering
 \includegraphics[width=0.7\linewidth]{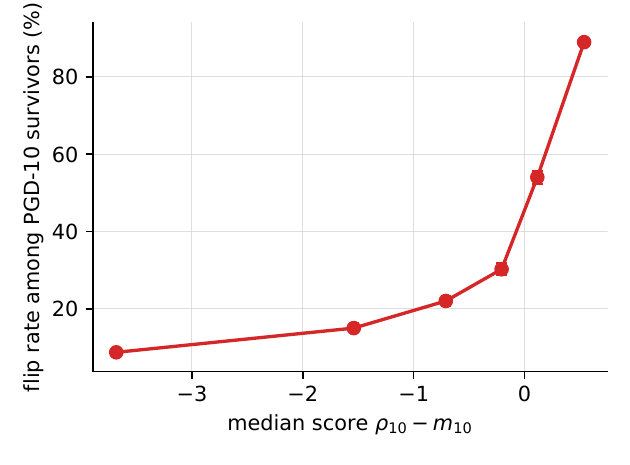}
 \caption{Per-sample flip rate for the score $\rho_{10}-m_{10}$ among
 $\PGD{10}$ survivors. Points show observed pass-$10$ to fail-$200$ flip
 rates within equal-count score bins, plotted against the bin-median score.
 Error bars show 95\% binomial confidence intervals. The score reaches pooled
 AUC $0.824$ for predicting pass-$10$ to fail-$200$ flips.}
 \label{fig:sample_flip_calibration_appendix}
\end{figure}

We also test whether sample-level coordinates predict which $\PGD{10}$ survivors are punctured by $\PGD{200}$.
The target is $\mathbb{1}[\mathrm{pass}_{10}=1,\mathrm{fail}_{200}=1]$ among shallow-budget survivors.
Among the simple scores compared below, $\rho_{10}-m_{10}$ reaches pooled AUC $0.824$ on $21{,}636$ pass-$10$ samples and $7{,}896$ pass-$10$ to fail-$200$ flips.
Figure~\ref{fig:sample_flip_calibration_appendix} shows a binned flip-rate
curve for this sample-level flip score.

\begin{table}[!t]
 \centering
 \caption{Per-sample flip prediction for simple scores among $\PGD{10}$ survivors.}
\label{tab:flip_scores}
\AppTableFont
\begin{adjustbox}{max width=\linewidth}
\begin{tabular}{lrr}
\toprule
Score & pooled AUC & median within-encoder AUC \\
\midrule
$\rho_{10}-m_{10}$ & $0.824$ & $0.856$ \\
raw FGSM-PGD gap & $0.820$ & $0.831$ \\
$\kappa_{10}$ & $0.768$ & $0.642$ \\
$-m_{10}$ & $0.755$ & $0.851$ \\
$\rho_{10}$ & $0.745$ & $0.657$ \\
$-\lambda$ & $0.695$ & $0.770$ \\
$\tau_{10}$ & $0.480$ & $0.521$ \\
\bottomrule
\end{tabular}
\end{adjustbox}
\end{table}

\section{Intervention evaluation details}
\label{app:intervention}

\begin{figure}[!t]
 \centering
 \includegraphics[width=0.75\linewidth]{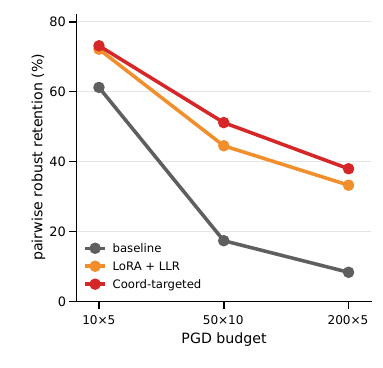}\\
 \includegraphics[width=0.75\linewidth]{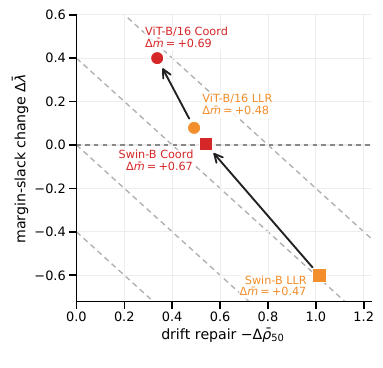}
 \caption{Coordinates separate drift repair from survival-margin repair.
	 \emph{(Top)} On ViT-B/16, both interventions improve pairwise robust
	 retention across $\PGD{10}\times5$, $\PGD{50}\times10$, and
	 $\PGD{200}\times5$, with stronger deep-budget retention for Coord.
	 \emph{(Bottom)} Component-median movements show that larger drift reduction can
	 yield a smaller survival-margin gain when margin slack is eroded. Dashed
	 diagonals are visual guides for combined slack-and-drift movement. Exact
	 $\Delta\bar m$ values are computed per sample before aggregation, with
	 $\Delta\bar\kappa$ outside this two-coordinate plane. Labeled
	 $\Delta\bar m$ values are median per-sample changes.}
 \label{fig:h2h}
\end{figure}

This section records the intervention cells used by the intervention analysis in Section~\ref{sec:intervention}.
Table~\ref{tab:cosface_vs_lora} summarizes the intervention cells that connect
margin-side repair, generic flattening, and coordinate-targeted repair, and
Figure~\ref{fig:h2h} plots the ViT-B/16 budget closure together with the
corresponding coordinate path.
The accompanying code and data artifact provides the per-cell records.

\paragraph{Training hyperparameters.}
Both LoRA interventions use rank $4$ adapters, AdamW at learning rate
$10^{-4}$, batch size $24$, one epoch over $5\%$ of the ImageNet-100 training
split, and a $10$-step inner attack. The coordinate-targeted objective runs its
inner attack under the evaluation threat model at $\ell_2$ $\eps=0.25$, while
local-linearity regularization uses its own perturbation scale of
$\ell_\infty$ $4/255$ for the gradient-agreement term. The coordinate-targeted
penalty weights are $\alpha=0.1$ and $\beta=\gamma=1$ with tolerances
$\eta_\kappa=\eta_\lambda=0.05$. The four reported seeds are $1$ through $4$.
Encoder parameters outside the adapters and the linear head stay frozen.

\paragraph{Matched evaluation subsets.}
Intervention effects are computed on matched baseline--intervention subsets.
For Swin-B, the baseline, LLR, and Coord evaluations share a common set of
$952$ clean-correct samples. For ViT-B/16, each method--seed pair uses its own
matched subset of $957$ to $967$ samples, and the baseline is recomputed on
that same subset before forming $\Delta R_{50}$ and the coordinate changes.
The corresponding baseline $R_{50}$ values range from $17.08\%$ to $17.19\%$
across these subsets. All reported intervention deltas are therefore
within-subset changes.

\begin{table*}[!t]
 \centering
 \AppTableFont
 \caption{Intervention classes read through the survival-margin coordinates. Margin-head retraining deltas are $n = 13$ encoder medians from the intervention-construction pool. LoRA + LLR and Coord deltas are four-seed means on ViT-B/16 at matched LoRA rank, optimizer, training budget, and PGD-$50\times10$ evaluation.}
 \label{tab:cosface_vs_lora}
 \begin{adjustbox}{max width=\linewidth}
 \begin{tabular}{lrrrrrl}
    \toprule
    Intervention & $\Delta\bar\lambda$ & $\Delta\bar\kappa$ & $\Delta\bar\rho$ & $\Delta\bar m$ & $\Delta R_{50}$ & Repairs \\
    \midrule
    Margin head retrain & $+0.396$ & $-0.043$ & $-0.161$ & $+0.514$ & $+10.3$\,pp & margin \\
    LoRA + LLR on ViT-B/16 & $+0.079$ & $-0.085$ & $-0.491$ & $+0.485$ & $+27.9$\,pp & drift via flattening \\
    \textbf{Coord on ViT-B/16} & $\mathbf{+0.400}$ & $\mathbf{-0.053}$ & $\mathbf{-0.337}$ & $\mathbf{+0.687}$ & $\mathbf{+33.3}$\,pp & survival margin \\
    \bottomrule
  \end{tabular}
 \end{adjustbox}
\end{table*}

\subsection{ViT-B/16 closure at PGD-200}
\label{app:vit_pgd200_closure}

\begin{table*}[!t]
 \centering
 \caption{Per-seed ViT-B/16 pairwise robust retention at the three evaluation budgets.}
 \label{tab:vit_seeds}
\AppTableFont
\begin{adjustbox}{max width=\linewidth}
\begin{tabular}{lccccc}
\toprule
Method & seed & $R_{10}$ & $R_{50}$ & $R_{200}$ & $\Delta R(10\to200)$ \\
\midrule
baseline & n/a & $61.20\%$ & $17.35\%$ & $8.31\%$ & $52.89$\,pp \\
LoRA + LLR & default & $72.66\%$ & $44.37\%$ & $31.13\%$ & $41.54$\,pp \\
LoRA + LLR & $1$ & $71.19\%$ & $46.79\%$ & $34.07\%$ & $37.12$\,pp \\
LoRA + LLR & $2$ & $73.19\%$ & $46.16\%$ & $34.28\%$ & $38.91$\,pp \\
LoRA + LLR & $3$ & $71.50\%$ & $44.69\%$ & $33.44\%$ & $38.07$\,pp \\
\midrule
LoRA + LLR & mean & $72.13 \pm 0.94\%$ & $44.50 \pm 1.04\%$ & $33.23 \pm 1.45\%$ & $38.91 \pm 1.90$\,pp \\
\midrule
Coord & $1$ & $72.66\%$ & $52.37\%$ & $38.07\%$ & $34.60$\,pp \\
Coord & $2$ & $73.40\%$ & $51.31\%$ & $38.38\%$ & $35.02$\,pp \\
Coord & $3$ & $73.29\%$ & $50.37\%$ & $37.12\%$ & $36.17$\,pp \\
Coord & $4$ & $73.08\%$ & $50.47\%$ & $38.17\%$ & $34.91$\,pp \\
\midrule
\textbf{Coord} & \textbf{mean} & $\mathbf{73.11 \pm 0.33\%}$ & $\mathbf{51.13 \pm 0.92\%}$ & $\mathbf{37.93 \pm 0.56\%}$ & $\mathbf{35.17 \pm 0.69}$\,pp \\
\bottomrule
\end{tabular}
\end{adjustbox}
\end{table*}

Paired bootstrap CIs on the per-sample collapse rate $\Pr(\mathrm{pass}_{10} \wedge \mathrm{fail}_{200})$ give $-14.01$\,pp $[-17.43,-10.54]$ for LoRA + LLR versus baseline, $-17.77$\,pp $[-21.50,-14.04]$ for Coord versus baseline, and $-3.73$\,pp $[-5.91,-1.58]$ for Coord versus LoRA + LLR. The intervention statistic $\Delta\bar m$ is the median of per-sample survival-margin changes. Component medians are reported as a coordinate path. The identity holds per sample before taking medians.

\subsection{Closure-fairness controls}
\label{app:closure_fairness}

\begin{table*}[!t]
 \centering
 \caption{Closure-fairness controls. Each row is one control condition at matched capacity, optimizer, training budget, and evaluation.}
 \label{tab:closure_fairness_tab}
\AppTableFont
\begin{adjustbox}{max width=\linewidth}
\begin{tabular}{lcccccc}
\toprule
Method & $\Delta\bar\lambda$ & $\Delta\bar\kappa$ & $\Delta\bar\rho$ & $\Delta\bar m$ & $\Delta R(10\to200)$ & collapse \\
\midrule
LLR, $\ell_\infty$ original & $+0.079 \pm 0.025$ & $-0.085 \pm 0.007$ & $-0.491 \pm 0.029$ & $+0.485 \pm 0.033$ & $38.9 \pm 1.9$\,pp & $38.9\%$ \\
LLR, $\ell_2$ threat-matched & $+0.350 \pm 0.004$ & $-0.052 \pm 0.006$ & $-0.238 \pm 0.028$ & $+0.536 \pm 0.027$ & $40.4 \pm 1.2$\,pp & $40.4\%$ \\
$\rho$-penalty ablation & $+0.018 \pm 0.007$ & $-0.003 \pm 0.001$ & $+0.015 \pm 0.015$ & $+0.000 \pm 0.010$ & $51.6 \pm 0.5$\,pp & $51.6\%$ \\
\textbf{Coord, full objective} & $\mathbf{+0.400 \pm 0.025}$ & $\mathbf{-0.053 \pm 0.004}$ & $\mathbf{-0.337 \pm 0.016}$ & $\mathbf{+0.687 \pm 0.018}$ & $\mathbf{35.2 \pm 0.7}$\,pp & $\mathbf{35.2\%}$ \\
\bottomrule
\end{tabular}%
\end{adjustbox}
\end{table*}

The $\ell_2$-matched LLR variant gives smaller survival-margin improvement than Coord. The $\rho$-penalty ablation stays at baseline collapse. The coordinate-targeted objective gives the closure effect.

\subsection{Guard-term ablation for the coordinate objective}
\label{app:coord_ablation}

The coordinate-targeted objective includes guard terms for $\kappa$ and $\lambda$. The four-seed head-to-head uses penalty weights $\alpha = 0.1$, $\beta = \gamma = 1$, and tolerances $\eta_\kappa = \eta_\lambda = 0.05$. The ablation below compares one guarded cell and one unguarded cell at the same LoRA capacity, optimizer, training budget, and evaluation as the four-seed head-to-head.

\begin{table}[!t]
 \centering
 \caption{Guard-term ablation. One guarded and one unguarded cell at matched LoRA capacity, optimizer, training budget, and evaluation.}
\label{tab:guard_ablation}
\AppTableFont
\begin{adjustbox}{max width=\linewidth}
\begin{tabular}{lcccc}
\toprule
Cell & $\Delta\bar\lambda$ & $\Delta\bar\kappa$ & $\Delta\bar\rho$ & $\Delta R_{50}$ \\
\midrule
guarded, $\beta = \gamma = 1$ & $+0.400$ & $-0.059$ & $-0.326$ & $+34.5$\,pp \\
unguarded, $\beta = \gamma = 0$ & $+0.310$ & $-0.048$ & $-0.321$ & $+31.8$\,pp \\
\bottomrule
\end{tabular}
\end{adjustbox}
\end{table}

The guard term is treated as an ablation check. Under the tested rank $4$ LoRA capacity, guarded and unguarded variants produce similar coordinate movement. The main comparison to LoRA + LLR uses the guarded four-seed cell.

\subsection{Swin-B head-to-head}
\label{app:swin_h2h}

\begin{table*}[!t]
 \centering
 \caption{Swin-B head-to-head coordinate deltas at $\PGD{50}\times10$.}
 \label{tab:swin_h2h_tab}
\AppTableFont
\begin{adjustbox}{max width=\linewidth}
\begin{tabular}{lcccc}
\toprule
Encoder and method & $\Delta\bar\lambda$ & $\Delta\bar\rho$ & $\Delta\bar m$ & Retention effect \\
\midrule
Swin-B, LLR & $-0.601$ & $-1.015$ & $+0.466$ & $+7.5$\,pp \\
Swin-B, Coord & $+0.003$ & $-0.541$ & $+0.666$ & $+25.7$\,pp \\
\bottomrule
\end{tabular}
\end{adjustbox}
\end{table*}

On Swin-B, generic flattening pays a large $\bar\lambda$ cost while
Coord holds $\bar\lambda$ near baseline, which gives the larger
retention gain.

\subsection{ViT-B/16 closure at PGD-500}
\label{app:pgd500_closure}

\begin{table*}[!t]
 \centering
 \caption{ViT-B/16 closure extended to $\PGD{500}$.}
 \label{tab:pgd500_closure_tab}
\AppTableFont
\begin{adjustbox}{max width=\linewidth}
\begin{tabular}{lcccccc}
\toprule
Cell & $R_{10}$ & $R_{200}$ & $R_{500}$ & $\Delta R(10\to200)$ & $\Delta R(10\to500)$ & collapse$_{500}$ \\
\midrule
baseline & $60.2\%$ & $8.2\%$ & $5.9\%$ & $+52.0$\,pp & $+54.3$\,pp & $54.8\%$ \\
LoRA + LLR, seed $1$ & $70.4\%$ & $33.7\%$ & $28.5\%$ & $+36.7$\,pp & $+41.9$\,pp & $42.1\%$ \\
Coord, seed $1$ & $71.5\%$ & $37.4\%$ & $32.5\%$ & $+34.0$\,pp & $+39.0$\,pp & $39.4\%$ \\
\bottomrule
\end{tabular}%
\end{adjustbox}
\end{table*}

Here collapse$_{500}$ denotes the per-sample pass-$10$ to fail-$500$ collapse
rate, $\Pr(\mathrm{pass}_{10}\wedge\mathrm{fail}_{500})$, paralleling the
collapse rate in Appendix~\ref{app:vit_pgd200_closure}. The $\PGD{500}$ check
uses one seed per method. The $\PGD{500}$ baseline uses the full primary split
rather than the intervention-construction subset, so $R_{10}$ and $R_{200}$
differ slightly from Appendix~\ref{app:vit_pgd200_closure}. The four-seed
$\PGD{200}$ comparison is the intervention measurement.

\section{Additional full-pool validation checks}
\label{app:additional_validation}

\subsection{Dynamic all-class coordinate recomputation}
\label{app:dynamic_allclass}

We recompute the coordinate record using the dynamic all-class margin
$G_{\mathrm{all}}(u)=z_y(u)-\max_{j\neq y}z_j(u)$.
We write $\bar\rho_{\mathrm{all},10}$ for the encoder-level median of $\rho$
computed under the dynamic all-class margin $G_{\mathrm{all}}$ at
$\PGD{10}\times5$. On the full $42$-encoder ImageNet-100 pool,
$\bar\rho_{\mathrm{all},10}$ ranks dynamic collapse from $\PGD{10}\times5$ to
$\PGD{200}\times5$ at Spearman $+0.797$ ($p=2.76\times10^{-10}$). The pairwise
and dynamic-all-class drift rankings are nearly identical, with Spearman
$+0.996$. This check indicates that the main ranking signal is not an artifact
of fixing the clean nearest competitor.

\subsection{Pairwise APGD solver variant}
\label{app:apgd_pairwise}

We rerun the fixed-competitor pairwise evaluation with an Auto-PGD (APGD) solver at
budgets $10\times5$ and $100\times5$. We write
$\bar\rho_{\mathrm{APGD},10}$ for the encoder-level median of $\rho$ measured
under the APGD solver at $10\times5$. On the full $42$-encoder pool,
$\bar\rho_{\mathrm{APGD},10}$ ranks APGD collapse at Spearman $+0.806$
($p=1.25\times10^{-10}$). The primary PGD and APGD drift rankings transfer at
Spearman $+0.980$. Thus the drift ranking is stable under this solver variant.

\subsection{Linear-head seed variability}
\label{app:head_seed}

We train three ImageNet-100 linear-head seeds on a representative $12$-backbone
subset and rerun the $\PGD{10}\times5$ coordinate evaluation. The seed-specific
$\bar\rho_{10}$ rankings agree with the primary ranking at Spearman $+0.993$,
$+1.000$, and $+1.000$. The median within-backbone standard deviation of
$\bar\rho_{10}$ across seeds is $0.00103$. Thus the drift ranking is stable to
linear-head seed variation in this subset.

\section{Standard-attack alignment checks}
\label{app:untargeted_ce}

We re-run untargeted cross-entropy APGD at $\ell_2$, $\eps = 0.25$ with step counts $\{10,100\}$.
On the full $42$-encoder pool, the encoder-level pairwise-margin $\bar\rho_{10}$ from the primary $\PGD{10}$ pipeline ranks the standard top-$1$ collapse $\Delta R_{\text{top1}}(10\to100)$ at Spearman $+0.833$, $p=8.1\times10^{-12}$.
The $\PGD{10}$ top-$1$ survival rate is uninformative for the same target, with Spearman $+0.061$.
The full $42$-encoder CE/APGD records are provided in the accompanying code and data artifact.
Here $R_{\text{untarg},k}$ denotes top-$1$ robust retention under the
untargeted CE/APGD attack with $k$ steps.
Table~\ref{tab:ce_apgd_subset} shows the original $11$-encoder subset used for the per-encoder records, where $\bar\rho_{10}$ ranks $\Delta R_{\text{untarg}}(\PGD{10}\to\PGD{100})$ at Spearman $+0.900$, $p = 1.6 \times 10^{-4}$.

\begin{table}[!t]
 \centering
 \caption{Original $11$-encoder subset used for the standard top-$1$ CE/APGD per-encoder records.}
\label{tab:ce_apgd_subset}
\AppTableFont
\begin{adjustbox}{max width=\linewidth}
\begin{tabular}{lcccc}
\toprule
Backbone & $\bar\rho_{10}^{\mathrm{pairwise}}$ & $R_{\text{untarg},10}$ & $R_{\text{untarg},100}$ & $\Delta R_{\text{untarg}}$ \\
\midrule
dino\_vits16 & $0.162$ & $81.6\%$ & $80.5\%$ & $+1.1$\,pp \\
efficientnet\_b0 & $0.185$ & $60.7\%$ & $58.0\%$ & $+2.7$\,pp \\
resnet18 & $0.288$ & $39.8\%$ & $38.1\%$ & $+1.7$\,pp \\
dino\_resnet50 & $0.352$ & $45.0\%$ & $32.2\%$ & $+12.9$\,pp \\
dinov2\_vitb14 & $0.354$ & $58.8\%$ & $48.5\%$ & $+10.3$\,pp \\
resnet50\_timm & $0.411$ & $48.7\%$ & $36.4\%$ & $+12.3$\,pp \\
deit3\_vitb16 & $0.435$ & $74.0\%$ & $55.7\%$ & $+18.2$\,pp \\
clip\_vitb16 & $0.487$ & $21.2\%$ & $7.8\%$ & $+13.4$\,pp \\
convnext\_small & $0.497$ & $55.5\%$ & $28.0\%$ & $+27.6$\,pp \\
deit\_tiny & $0.513$ & $35.0\%$ & $16.9\%$ & $+18.1$\,pp \\
convnext\_tiny & $0.566$ & $48.4\%$ & $22.1\%$ & $+26.3$\,pp \\
\bottomrule
\end{tabular}
\end{adjustbox}
\end{table}

\section{Representation-geometry correlation}
\label{app:nc}

We compute the within-class scatter ratio NC$1$ \citep{papyan2020prevalence} on each encoder's frozen ImageNet-100 validation features. Across the analytical $n = 40$ pool, smaller NC$1$ is associated with larger $\bar\kappa$ and $\bar\rho$ in tandem. Spearman$(\mathrm{NC}1,\bar\kappa) = -0.562$, $p = 0.0002$, and Spearman$(\mathrm{NC}1,\bar\rho) = -0.492$, $p = 0.0012$. The same correlation against the encoder-level median of $\tau_i=\rho_i-\kappa_i$ is $-0.133$, $p = 0.41$. The representation-level pattern is consistent with residual compression, since NC$1$ correlates with $\bar\kappa$ and $\bar\rho$ separately, while its correlation with their net residual $\bar\tau$ is small and statistically unresolved.

\section[L-infinity evaluation]{$\ell_\infty$ evaluation}
\label{app:linf}

The coordinate construction extends to $\ell_\infty$ through the
support-function form (Appendix~\ref{app:identity}). For this threat model, the
one-step perturbation is
\[
\delta_{\mathrm{FGSM}}=-\eps\,\mathrm{sign}(g),
\]
so the support-function scale is $\Lp=\eps\|g\|_1$. PGD candidates are
selected with the same best-drop fixed-competitor convention as in
Section~\ref{ssec:def}. The $\ell_\infty$ solver projects to the
$\ell_\infty$ ball after each step and clips inputs to $[0,1]$.
The step-size set remains $\{\eps/4,\eps/10\}$, and the restart convention
follows Appendix~\ref{app:convergence}. We run the full 42-encoder pool at
$\ell_\infty$ $\eps=0.375/255$ with $\PGD{10}\times5$ and $\PGD{100}\times5$.

\subsection{Primary result}
\label{app:linf_primary}

Shallow drift $\bar\rho_{10}^{\ell_\infty}$ ranks
$\ell_\infty$ cross-budget collapse
$\Delta R^{\ell_\infty}(\PGD{10}\times5\to\PGD{100}\times5)$ at
Spearman $+0.819$, $p=3.4\times10^{-11}$.

The largest $\ell_\infty$ collapses are at ViT-B/16 ($27.7$\,pp),
Swin-Small ($27.0$\,pp), and Swin-Base ($25.7$\,pp). Low-collapse
encoders include DINO ViT-B/16 ($1.7$\,pp), EfficientNet-B0
($1.8$\,pp), and DINO ViT-S/16 ($2.8$\,pp), matching the $\ell_2$
low-collapse encoders.

\subsection{Cross-norm rank consistency}
\label{app:linf_crossnorm}

On the same 42-encoder pool,
$\mathrm{Spearman}(\bar\rho_{10}^{\ell_2},
\bar\rho_{10}^{\ell_\infty})=+0.972$ ($p=1.03\times10^{-26}$, 95\%
CI $[+0.927,+0.987]$) at the $\ell_\infty$ operating radius
$\eps=0.375/255$. Table~\ref{tab:cross_norm} reports cross-norm rank consistency
across $\ell_\infty$ radii, with $\bar\rho_{10}^{\ell_2}$ fixed at
$\ell_2$ $\eps=0.25$.

\begin{table}[!t]
 \centering
 \caption{Cross-norm rank consistency across $\ell_\infty$ radii, with $\bar\rho_{10}^{\ell_2}$ fixed at $\ell_2$ $\eps=0.25$.}
\label{tab:cross_norm}
\AppTableFont
\begin{adjustbox}{max width=\linewidth}
\begin{tabular}{cccc}
\toprule
$\eps_{\ell_\infty}$ (in $1/255$) & $n$ &
$\mathrm{Spearman}(\bar\rho_{10}^{\ell_2},\bar\rho_{10}^{\ell_\infty})$
& $p$ \\
\midrule
$0.25$ & $42$ & $+0.903$ & $3.1\times10^{-16}$ \\
$\mathbf{0.375}$ & $\mathbf{42}$ & $\mathbf{+0.972}$
& $\mathbf{1.0\times10^{-26}}$ \\
$0.5$ & $42$ & $+0.933$ & $2.4\times10^{-19}$ \\
$0.75$ & $42$ & $+0.810$ & $8.3\times10^{-11}$ \\
$1.0$ & $42$ & $+0.623$ & $1.1\times10^{-5}$ \\
\bottomrule
\end{tabular}
\end{adjustbox}
\end{table}

Cross-norm rank consistency exceeds $+0.90$ across
$\eps_{\ell_\infty}\in\{0.25,0.375,0.5\}/255$ and remains positive at
all tested radii. The attenuation at $\eps\ge0.75/255$ follows
retention floor compression (median $R_{10}^{\ell_\infty}$ near $15\%$ or
below).

\subsection{Radius sweep}
\label{app:linf_sweep}

The $\ell_\infty$ within-norm collapse ranking also varies with radius.
Table~\ref{tab:linf_radius_sweep} reports median retention and the Spearman
correlation between $\bar\rho_{10}^{\ell_\infty}$ and
$\Delta R^{\ell_\infty}(\PGD{10}\times5\to\PGD{100}\times5)$ for
each radius.

\begin{table}[!t]
 \centering
 \caption{Within-norm $\ell_\infty$ radius sweep. Median retention and the rank correlation with within-norm collapse at each radius.}
\label{tab:linf_radius_sweep}
\AppTableFont
\begin{adjustbox}{max width=\linewidth}
\begin{tabular}{ccccc}
\toprule
$\eps$ (in $1/255$) & median $R_{10}$ & median $R_{100}$
& median collapse & Spearman \\
\midrule
$0.25$ & $66.69\%$ & $56.05\%$ & $6.81$\,pp & $+0.842$ \\
$\mathbf{0.375}$ & $\mathbf{48.71\%}$ & $\mathbf{33.89\%}$
& $\mathbf{9.75}$\,pp & $\mathbf{+0.819}$ \\
$0.5$ & $33.35\%$ & $17.80\%$ & $10.66$\,pp & $+0.696$ \\
$0.75$ & $15.13\%$ & $4.68\%$ & $6.40$\,pp & $+0.447$ \\
$1.0$ & $6.15\%$ & $1.32\%$ & $3.75$\,pp & $+0.388$ \\
\bottomrule
\end{tabular}
\end{adjustbox}
\end{table}

The within-norm Spearman declines monotonically with $\eps$, following
retention floor compression.
At $\eps=0.375/255$, median $R_{10}^{\ell_\infty}=48.7\%$, giving
collapse variation comparable to the primary $\ell_2$ regime.

\subsection[Cancellation regime at L-infinity]{Cancellation regime at $\ell_\infty$}
\label{app:linf_cancellation}

At $\ell_\infty$ $\eps=1/255$, encoder-level coordinates show the
cancellation pattern described in Section~\ref{sec:decomposition}:

\begin{table}[!t]
 \centering
 \caption{Encoder-level coordinates at $\ell_\infty$ $\eps=1/255$ showing the cancellation pattern.}
\label{tab:linf_cancellation}
\AppTableFont
\begin{adjustbox}{max width=\linewidth}
\begin{tabular}{lrrrl}
\toprule
Encoder & $\bar\tau$ & $\bar\kappa$ & $\bar\rho$ & regime \\
\midrule
DINO ViT-S/16 & $0.51$ & $0.08$ & $0.60$ & drift \\
ResNet-50 & $0.23$ & $0.92$ & $1.15$ & cancellation \\
CLIP ViT-B/16 & $0.07$ & $0.87$ & $0.94$ & cancellation \\
\bottomrule
\end{tabular}
\end{adjustbox}
\end{table}

ResNet-50 and CLIP show low $\bar\tau$ produced by offsetting high
$\bar\kappa$ and high $\bar\rho$, instantiating the cancellation
regime in a different threat model.

\subsection[Attack adequacy at L-infinity]{Attack adequacy at $\ell_\infty$}
\label{app:linf_attack}

AutoAttack targets top-$1$ misclassification, whereas our evaluation
optimizes the fixed-competitor pairwise margin, so we use this comparison as
an attack-adequacy sanity check for the recorded pairwise objective.
As a high-radius stress check, on 8 representative encoders at $\ell_\infty$
$\eps=1/255$,
fixed-competitor pairwise PGD finds a deeper margin drop than
untargeted AutoAttack on $68$--$89\%$ of clean-correct samples.
The comparison indicates that the fixed-competitor
pairwise-margin record is not dominated by a standard untargeted attack under
the same threat model.